\documentclass[10pt,journal,compsoc]{IEEEtran}
\usepackage{times}
\usepackage{epsfig}
\usepackage{graphicx}
\usepackage{amsmath}
\usepackage{amssymb}
\usepackage{booktabs}
\usepackage{subcaption}
\usepackage{adjustbox}
\usepackage{comment}

\usepackage{todonotes}
\usepackage{pifont}
\usepackage{multirow}
\usepackage[numbers,sort&compress]{natbib}
\usepackage{enumitem}
\usepackage{todonotes}
\usepackage{courier}
\usepackage{amssymb}

\usepackage{mathtools}
\usepackage{tablefootnote}
\usepackage[bottom]{footmisc}

\definecolor{start}{RGB}{255,255,255}
\definecolor{c1}{RGB}{51,160,44}
\definecolor{c2}{RGB}{55,126,184} %
\definecolor{c3}{RGB}{255, 255, 255}
\definecolor{c4}{RGB}{228,26,28}
\definecolor{c5}{RGB}{255,127,0}

\newcommand{\arch}{\texttt{DiCENet}}
\newcommand{\dice}{\texttt{DiCE}}
\newcommand{\dimconv}{\texttt{DimConv}}
\newcommand{\ecf}{\texttt{DimFuse}}

\definecolor{gg}{RGB}{44,162,95}
\definecolor{highlightArch}{RGB}{0,0,0}

\providecommand{\redF}[1]{{\protect\color{black}{\bf -#1}}}
\providecommand{\incF}[1]{{\protect\color{black}{+#1}}}
\providecommand{\incAcc}[1]{{\protect\color{black}{\bf +#1}}}

\definecolor{depthwise}{rgb}{0.0, 1.0, 1.0}
\definecolor{widthwise}{rgb}{0.62, 0.0, 0.77}
\definecolor{heightwise}{rgb}{0.0, 0.5, 0.0}

\usepackage{tikz-3dplot}
\usepackage{tikz}
\usetikzlibrary{decorations.pathreplacing, calligraphy}
\usetikzlibrary{shapes.callouts}
\usetikzlibrary{chains, fit, quotes, automata}
\usetikzlibrary{arrows}
\usetikzlibrary{matrix,positioning}
\usetikzlibrary{shapes.geometric, shapes.misc}
\usetikzlibrary{calc}
\usepackage[bookmarks=false]{hyperref}

\makeatletter
\long\def\@IEEEtitleabstractindextextbox#1{\parbox{0.922\textwidth}{#1}}
\makeatother

\begin{document}
%
\title{\arch: Dimension-wise Convolutions for Efficient Networks}
%
%
%
%

\author{Sachin~Mehta, Hannaneh~Hajishirzi, and~Mohammad~Rastegari 
\IEEEcompsocitemizethanks{\IEEEcompsocthanksitem S. Mehta, H. Hajishirzi, and M. Rastegari are with the University of Washington, Seattle, WA, 98195.\protect\\
E-mail: \{sacmehta, hannaneh, mrast\}@cs.washington.edu}}
\IEEEtitleabstractindextext{%
\begin{abstract}
 We introduce a novel and generic convolutional unit, \dice~unit, that is built using dimension-wise convolutions and dimension-wise fusion. The dimension-wise convolutions apply light-weight convolutional filtering across each dimension of the input tensor while dimension-wise fusion efficiently combines these dimension-wise representations; allowing the \dice~unit to efficiently encode spatial and channel-wise information contained in the input tensor. The \dice~unit is simple and can be seamlessly integrated with any architecture to improve its efficiency and performance. Compared to depth-wise separable convolutions, the \dice~unit shows significant improvements across different architectures. When \dice~units are stacked to build the \arch~model, we observe significant improvements over state-of-the-art models across various computer vision tasks including image classification, object detection, and semantic segmentation. On the ImageNet dataset, the \arch~delivers 2-4\% higher accuracy than state-of-the-art manually designed models (e.g., MobileNetv2 and ShuffleNetv2). Also, \arch~generalizes better to tasks (e.g., object detection) that are often used in resource-constrained devices in comparison to state-of-the-art separable convolution-based efficient networks, including neural search-based methods (e.g., MobileNetv3 and MixNet). %
\end{abstract}

\begin{IEEEkeywords}
Convolutional Neural Network, Image Classification, Object Detection, Semantic Segmentation, Efficient Networks.
\end{IEEEkeywords}}

\maketitle

\IEEEdisplaynontitleabstractindextext

%
\IEEEpeerreviewmaketitle

\begin{figure}[t!]
    \centering
    \begin{subfigure}[b]{\columnwidth}
        \centering
        \includegraphics[width=0.9\columnwidth]{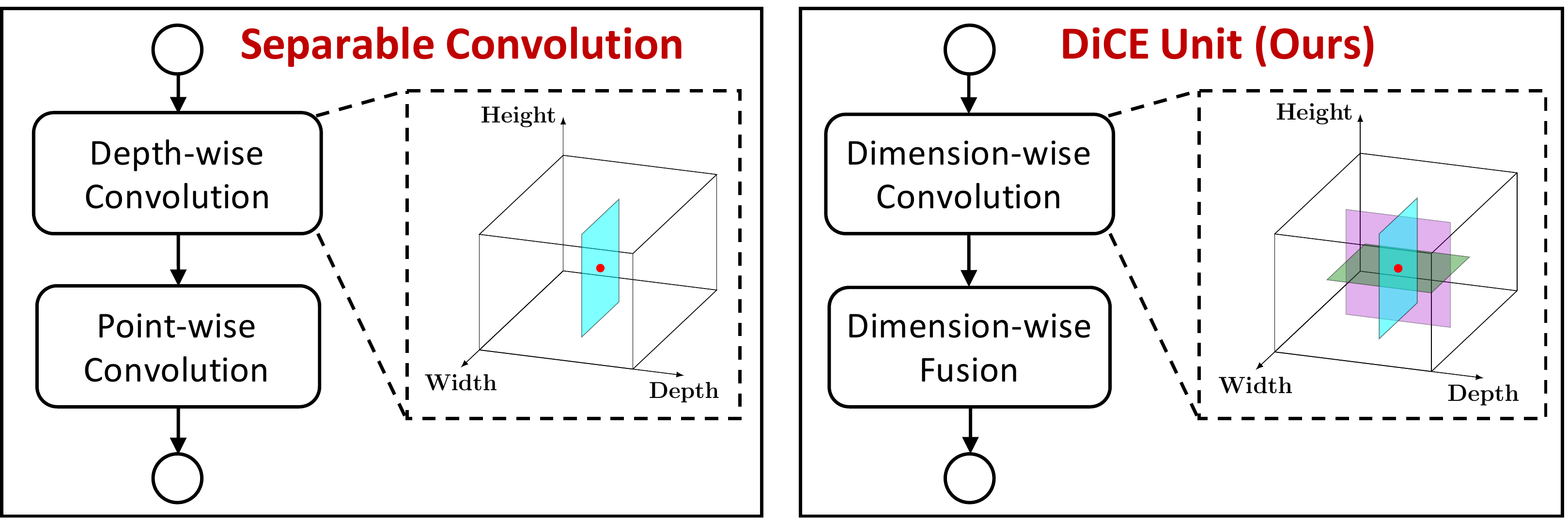}
        \caption{Block of separable convolutions and the \dice~unit. Convolutional kernels are highlighted in color (\textbf{\textcolor{depthwise}{depth-}}, \textbf{\textcolor{widthwise}{width-}}, and \textbf{\textcolor{heightwise}{height-}}wise).}
        \label{fig:depthVsDimVis}
    \end{subfigure}
    \vfill
    \begin{subfigure}[b]{\columnwidth}
        \centering
        \includegraphics[width=0.85\columnwidth]{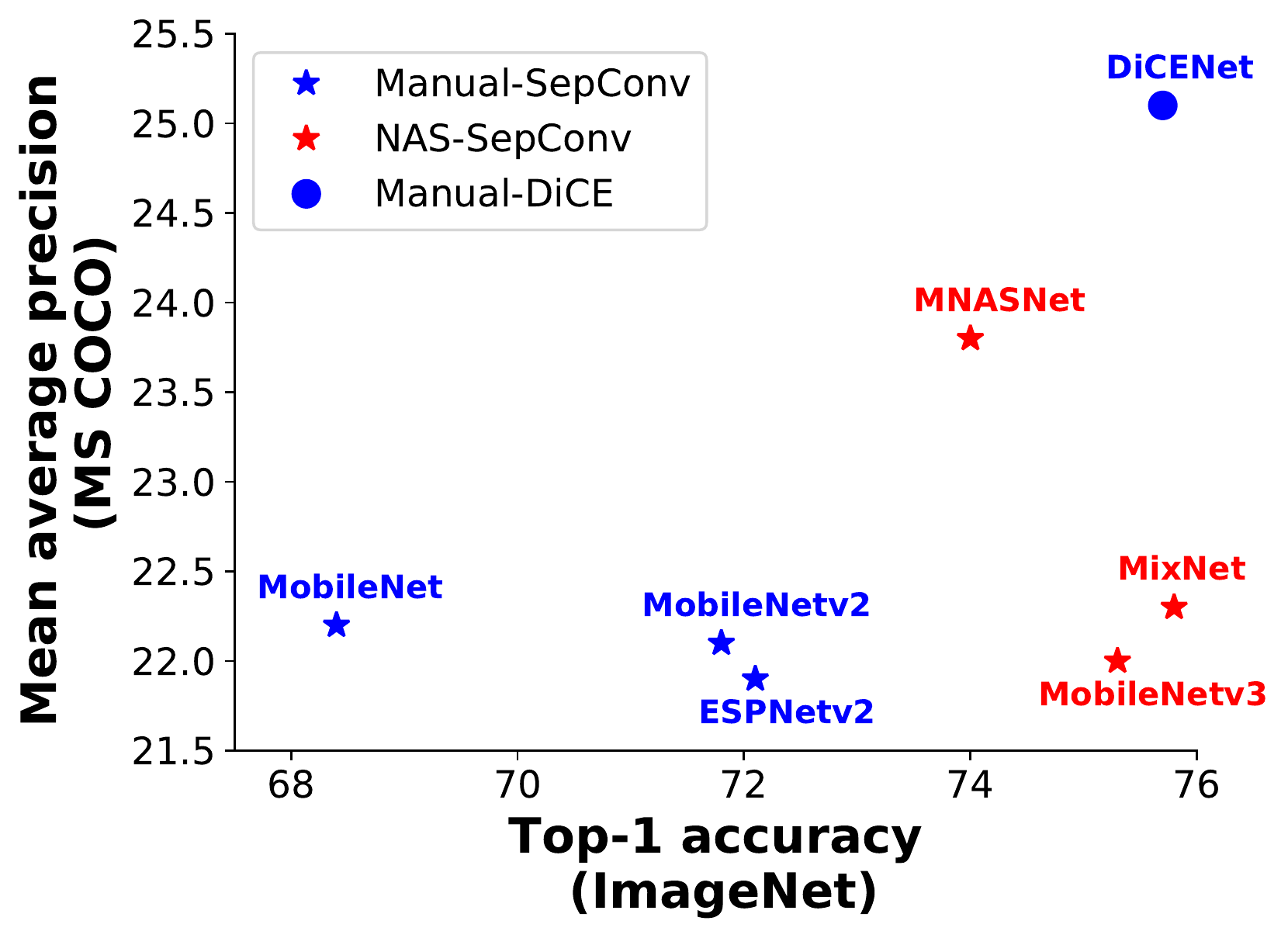}
        \caption{Network with the \dice~unit (\arch) has better \textbf{task-level generalization} properties compared to networks with separable convolutions (MobileNet \cite{howard2017mobilenets}, MobileNetv2 \cite{sandler2018mobilenetv2}, \textcolor{black}{MobileNetv3} \cite{howard2019searching}, \textcolor{black}{MixNet} \cite{Tan2019MixConvMD}, \textcolor{black}{MNASNet} \cite{tan2018mnasnet}). Here, Manual-SepConv and NAS-SepConv represents the models that uses separable convolution without and with neural architecture search (NAS), respectively. Manual-DiCE represents the \arch~without NAS. On the ImageNet dataset, these networks have about 200-300 million floating point operations. See Section \ref{sec:resultsImg} for more details.}
        \label{fig:taskGen}
    \end{subfigure}
    \caption{\bfseries Separable convolutions vs. the \dice~unit.}
    \label{fig:compareFlops}
\end{figure}

\IEEEraisesectionheading{\section{Introduction}
\label{sec:intro}}
\IEEEPARstart{T}{he} basic building layer at the heart of convolutional neural networks (CNNs) is a convolutional layer that encodes spatial and channel-wise information simultaneously \cite{krizhevsky2012imagenet,simonyan2014very,he2016deep}. Learning representations using this layer is computationally expensive. Improving the efficiency of CNN architectures as well as convolutional layers is an active area of research. Most recent attempts have focused on improving the efficiency of CNN architectures using compression- and quantization-based methods (e.g. \cite{li2018constrained,he2018amc}). Recently, several factorization-based methods have been proposed to improve the efficiency of standard convolutional layers (e.g. \cite{jin2014flattened, howard2017mobilenets, mehta2018espnet}). In particular, depth-wise separable convolutions \cite{howard2017mobilenets, chollet2017xception} have gained a lot of attention (see Figure \ref{fig:depthVsDimVis}). These convolutions have been used in several efficient state-of-the-art architectures, including neural search-based architectures \cite{sandler2018mobilenetv2, mehta2018espnetv2, tan2018mnasnet, wu2018fbnet}.

Separable convolutions factorize the standard convolutional layer in two steps: (1) a light-weight convolutional filter is applied  to each input channel using depth-wise convolutions \cite{chollet2017xception} to learn spatial representations, and (2) a point-wise ($1\times1$) convolution is applied to learn linear combinations between spatial representations. Though depth-wise convolutions are efficient, they do not encode channel-wise relationships. Therefore, separable convolutions rely on point-wise convolutions to encode channel-wise relationships. This puts a significant computational load on point-wise convolutions and makes them a computational bottleneck. For example, point-wise convolutions account for about 90\% of total operations in ShuffleNetv2 \cite{ma2018shufflenet} and MobileNetv2 \cite{sandler2018mobilenetv2}.

In this paper, we introduce \arch, \textbf{Di}mension-wise \textbf{C}onvolutions for \textbf{E}fficient \textbf{Net}works that encodes spatial and channel-wise representations efficiently. Our main contribution is the novel and generic module, the \dice~unit (Figure \ref{fig:depthVsDimVis}), that is built using \texttt{Dim}ension-wise \texttt{Conv}olutions (\dimconv) and \texttt{Dim}ension-wise \texttt{Fus}ion (\ecf).  The \dimconv~applies a light-weight convolutional filter across ``each dimension" of the input tensor to learn local dimension-wise representations while \ecf~efficiently combines these dimension-wise representations to incorporate global information. 

With \dimconv~and \ecf, we build  an efficient convolutional unit, the \dice~unit, that can be easily integrated into existing or new CNN architectures to improve their performance and efficiency. Figure \ref{fig:compareFlops} shows that the \dice~unit is effective in comparison to widely-used separable convolutions. Compared to state-of-the-art manually designed networks (e.g., MobileNet \cite{howard2017mobilenets}, MobileNetv2 \cite{sandler2018mobilenetv2}, and ShuffleNetv2 \cite{ma2018shufflenet}), \arch~delivers significantly better performance. For example, for a network with about 300 million floating point operations (FLOPs), \arch~is about 4\% more accurate than MobileNetv2. Importantly, \arch, a manually designed network, delivers similar or better performance than neural architecture search (NAS) based methods. For example, \arch~is 1.7\% more accurate than MNASNet for a network with about 300 MFLOPs.

We empirically demonstrate in Section \ref{sec:resultsImg} and Section \ref{sec:results} that the \arch~network, built by stacking \dice~units, achieves significant improvements on standard benchmarks across different tasks over existing networks. Compared to existing efficient networks, \arch~generalizes better to tasks (e.g., object detection) that are often used in resource-constrained devices. For instance, \arch~achieves about 3\% higher mean average precision than MobileNetv3 \cite{howard2019searching} and MixNet \cite{Tan2019MixConvMD} on the MS-COCO object detection task with SSD \cite{liu2016ssd} as a detection pipeline (Fig. \ref{fig:taskGen}). Our source code in PyTorch is open-source and is available at \url{https://github.com/sacmehta/EdgeNets/}.

\section{Related Work}

\noindent \textbf{CNN architecture designs:} Recent success in visual recognition tasks, including classification, detection, and segmentation, can be attributed to the exploration of different CNN designs (e.g., \cite{krizhevsky2012imagenet,simonyan2014very,he2016deep,howard2017mobilenets}). The basic building layer in these networks is a standard convolutional layer, which is computationally expensive. Factorization-based methods improve the efficiency of these layers. For instance, flattened convolutions \cite{jin2014flattened} approximate a standard convolutional layer with three point-wise convolutions that are applied sequentially, one point-wise convolution per tensor dimension. These convolutions ignore spatial relationships between pixels and do not generalize across a wide variety of computer vision tasks (e.g., detection and segmentation) and large scale datasets (e.g., ImageNet and MS-COCO). To improve the efficiency of standard convolutions while maintaining the performance and generalization ability at scale, depth-wise separable convolutions \cite{howard2017mobilenets} are proposed that factorizes the standard convolutional layer into depth-wise and point-wise convolution layers. Most of the efficient CNN architectures are built using these separable convolutions, including MobileNets \cite{howard2017mobilenets, sandler2018mobilenetv2}, ShuffleNets \cite{zhang2017shufflenet, ma2018shufflenet}, and ESPNetv2 \cite{mehta2018espnetv2}. In this work, we introduce dimension-wise convolutions that generalize depth-wise convolutions to all dimensions of the input tensor. We also introduce an efficient way for combining these dimension-wise representations. As confirmed by our experiments in Section \ref{sec:genEff}, \ref{sec:resultsImg} and \ref{sec:results}, the \dice~unit is more effective than separable convolutions.

\vspace{1mm}
\noindent \textbf{Neural architecture search:} Recently, neural architecture search-based methods have been proposed to automatically construct network architectures (e.g., \cite{zoph2016neural,wu2018fbnet,zoph2018learning,tan2018mnasnet, howard2019searching}). These methods search over a large network space (e.g., MNASNet \cite{tan2018mnasnet} searches over 8K different design choices) using a dictionary of pre-defined search space parameters, including different types of convolutional layers and kernel sizes, to find a heterogeneous network structure that satisfies optimization constraints, such as inference time. The proposed unit is novel and cannot be discovered using existing neural search-based methods (e.g., \cite{tan2018mnasnet, wu2018fbnet, Wortsman2019DiscoveringNW}). However, we believe that better neural architectures can be discovered by adding the \dice~unit in neural search dictionary.

\vspace{1mm}
\noindent \textbf{Quantization, compression, and distillation:} Network quantization-based approaches \cite{rastegari2016xnor,wu2016quantized,courbariaux2016binarized,andri2018yodann} approximate 32-bit full precision convolution operations with fewer bits. This improves inference speed and reduces the amount of memory required for storing network weights.  Network compression-based approaches \cite{han2015deep,wen2016learning,li2018constrained,veit2018convolutional,he2018amc} improve the efficiency of a network by removing redundant weights and connections. Unlike network quantization and compression, distillation-based approaches \cite{hinton2015distilling,gupta2016cross,yim2017gift} improve the accuracy of (usually shallow) networks by supervising the training with large pre-trained networks. These approaches are effective for improving the efficiency of a network, including efficient architecture designs (e.g., \cite{jacob2018quantization, wang2019haq}) and are orthogonal to our work. We believe that the efficiency of \arch~can be further improved using these methods.
\section{\arch}
\label{sec:model}
\begin{figure}[b!]
    \centering
    \includegraphics[width=0.75\columnwidth]{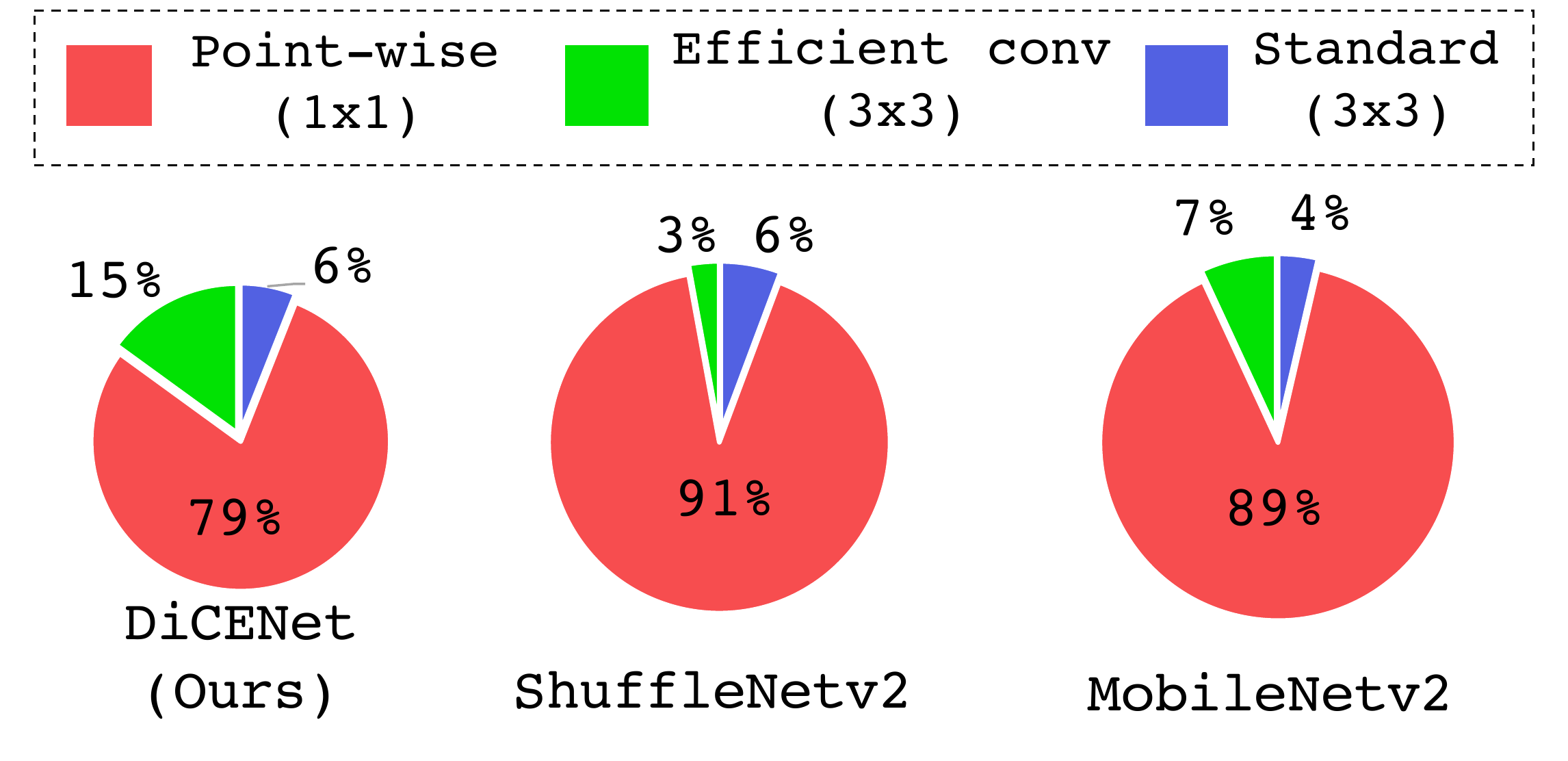}
    \caption{\textbf{Convolution-wise distribution of FLOPs} for different networks with similar accuracy. The size of pie charts is scaled with respect to MobileNetv2's FLOPs. In \arch, efficient conv's correspond to dimension-wise convolutions while in other networks, they correspond to depth-wise convolutions.}
    \label{fig:convDist}
\end{figure}

Standard convolutions encode spatial and channel-wise information \textit{simultaneously}, but they are computationally expensive. To improve the efficiency of standard convolutions, separable (or depth-wise separable) convolutions are introduced \cite{howard2017mobilenets}, where spatial and channel-wise information is encoded \textit{separately} using depth-wise and point-wise convolutions, respectively. Though this factorization is effective, it puts a significant computational load on point-wise convolutions and makes them a computational bottleneck (see Figure \ref{fig:convDist}).

\begin{figure*}[t!]
    \centering
    \includegraphics[width=1.8\columnwidth]{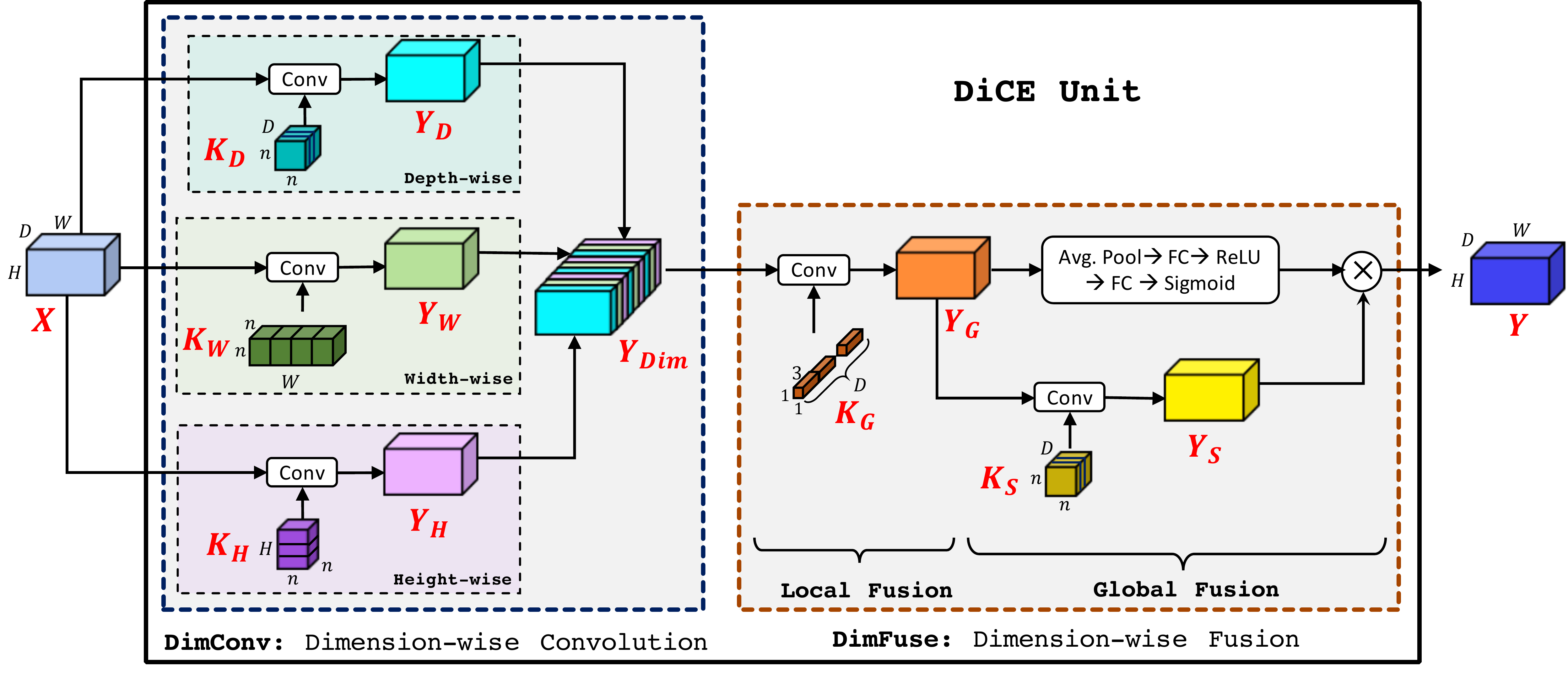}
    \caption{\textbf{\dice~unit} efficiently encodes the spatial and channel-wise information in the input tensor $\mathbf{X}$ using dimension-wise convolutions (\dimconv) and dimension-wise fusion (\ecf) to produce an output tensor $\mathbf{Y}$. For simplicity, we show kernel corresponding to each dimension independently. However, in practice, these three kernels are executed simultaneously, leading to faster run-time. See Section \ref{ssec:archDetails} and \ref{sec:resultsImg} for more details.}
    \label{fig:dwconv}
\end{figure*}

To encode spatial and channel-wise information efficiently, we introduce the \dice~unit and is shown in Figure \ref{fig:dwconv}. The \dice~unit factorizes standard convolution using \texttt{Dim}ension-wise \texttt{Conv}olution (\dimconv, Section \ref{ssec:dimConv}) and \texttt{Dim}ension-wise \texttt{Fus}ion (\ecf, Section \ref{ssec:dimFuse}). \dimconv~applies a light-weight filtering across each dimension of the input tensor to learn local dimension-wise representations. \ecf~efficiently combines these representations from different dimensions and incorporates global information. The ability to encode local spatial and channel-wise information from all dimensions using \dimconv~enables the \dice~unit to use \ecf~instead of computationally expensive point-wise convolutions.

\subsection{Dimension-wise Convolution (\dimconv)}
\label{ssec:dimConv}
We use dimension-wise convolutions (\dimconv) to encode depth-, width-, and height-wise information \textit{independently}. To achieve this, \dimconv~extends depth-wise convolutions to \textit{all dimensions} of the input tensor $\mathbf{X} \in \mathbb{R}^{D \times H \times W}$, where $W$, $H$, and $D$ corresponds to width, height, and depth of $\mathbf{X}$. As illustrated in Figure \ref{fig:dwconv}, \dimconv~has three branches, one branch per dimension. These branches apply $D$ depth-wise convolutional kernels $\mathbf{k}_{D} \in \mathbb{R}^{1 \times n \times n}$ along depth, $W$ width-wise convolutional kernels $\mathbf{k}_{W} \in \mathbb{R}^{n \times n \times 1}$ along width, and $H$ height-wise convolutional kernels $\mathbf{k}_{H} \in \mathbb{R}^{n \times 1 \times n}$ kernels along height to produce outputs $\mathbf{Y_{D}}$, $\mathbf{Y_{W}}$, and $\mathbf{Y_{H}} \in \mathbb{R}^{D \times H \times W}$ that encode information from all dimensions of the input tensor. The outputs of these independent branches are concatenated along the depth dimension, such that the first spatial plane of $\mathbf{Y_{D}}$, $\mathbf{Y_{W}}$, and $\mathbf{Y_{H}}$ are put together and so on, to produce the output $\mathbf{Y_{Dim}} = \{\mathbf{Y_{D}}, \mathbf{Y_{W}}, \mathbf{Y_{H}}\} \in \mathbb{R}^{3D \times H \times W}$.

\subsection{Dimension-wise Fusion (\ecf)} 
\label{ssec:dimFuse}
The dimension-wise convolutions encode local information from different dimensions of the input tensor, but do not capture global information. A standard approach to combine features globally in CNNs is to use a point-wise convolution \cite{he2016deep, howard2017mobilenets}. A point-wise convolutional layer applies $D$ point-wise kernels $\mathbf{k}_p \in \mathbb{R}^{3D \times 1 \times 1}$ and performs $3D^2HW$ operations to combine dimension-wise representations of $\mathbf{Y_{Dim}} \in \mathbb{R}^{3D \times H \times W}$ and produce an output $\mathbf{Y} \in \mathbb{R}^{D \times H \times W}$. This is computationally expensive. Given the ability of \dimconv~to encode spatial and channel-wise information (though independently), we introduce a fusion module, Dimension-wise fusion (\ecf), that allows us to combine representations of $\mathbf{Y_{Dim}}$ efficiently. As illustrated in Figure \ref{fig:dwconv}, \ecf~factorizes the point-wise convolution in two steps: (1) local fusion and (2) global fusion. 

$\mathbf{Y_{Dim}} \in \mathbb{R}^{3D \times H \times W}$ concatenates spatial planes along depth dimension from $\mathbf{Y_{D}}$, $\mathbf{Y_{W}}$, and $\mathbf{Y_{H}}$ (see Figure \ref{fig:dwconv}). Therefore, $\mathbf{Y_{Dim}}$ can be viewed as a tensor with $D$ groups, each group with three spatial planes (one from each dimension). \ecf~uses a group point-wise convolutional layer to combine dimension-wise information contained in $\mathbf{Y_{Dim}}$. In particular, this group convolutional layer applies $D$ point-wise convolutional kernels $\mathbf{k}_G \in \mathbb{R}^{3 \times 1 \times 1}$ to $\mathbf{Y_{Dim}} $ and produces an output $\mathbf{Y_{G}} \in \mathbb{R}^{D \times H \times W}$. Since $D$ kernels in $\mathbf{k}_G$ operates independently on $D$ groups in $\mathbf{Y_{Dim}}$, we call this {\it local } fusion operation.
 
To efficiently encode the global information in $\mathbf{Y_{G}}$, \ecf~learns spatial and channel-wise representations independently and then propagate channel-wise encodings to spatial encodings using an element-wise multiplication. Specifically, \ecf~encodes spatial representations by applying $D$ depth-wise convolutional kernels $\mathbf{k}_{S} \in \mathbb{R}^{1 \times n \times n}$ to $\mathbf{Y_{G}}$ to produce an output $\mathbf{Y}_{S}$.\footnote{When the depth of $\mathbf{Y_{Dim}}$ is different from $\mathbf{Y}$, then $\mathbf{k}_{S}$ is a group convolution, where number of groups is the greatest common divisor between the depth of $\mathbf{Y_{Dim}}$ and $\mathbf{Y}$.} Motivated by Squeeze-Excitation (SE) unit \cite{hu2018squeeze}, we squeeze spatial dimensions of $\mathbf{Y_{G}}$ and encode channel-wise representations using two fully connected (FC) layers. The first FC layer reduces the input dimension from $D$ to $\frac{D}{4}$ while the second FC layer expands dimensionality from $\frac{D}{4}$ to $D$. To allow these fully connected layers to learn non-linear representations, a ReLU activation is added in between these two layers. Similar to the SE unit, spatial representations $\mathbf{Y_G}$ are then scaled using these channel-wise representations to produce output $\mathbf{Y}$.

The computational cost of \ecf~is $HWD(3 + n^2 + D)$. Effectively, \ecf~reduces the computational cost of point-wise convolutions by a factor of $\frac{3D}{3 + n^2 + D}$. \ecf~uses $n=3$, so the computational cost is approximately $3\times$ smaller than that of the point-wise convolution.

\definecolor{boxA}{RGB}{169, 209, 142}
\definecolor{boxB}{RGB}{244, 177, 131}
\begin{figure*}[t!]
    \centering
    \includegraphics[height=140px]{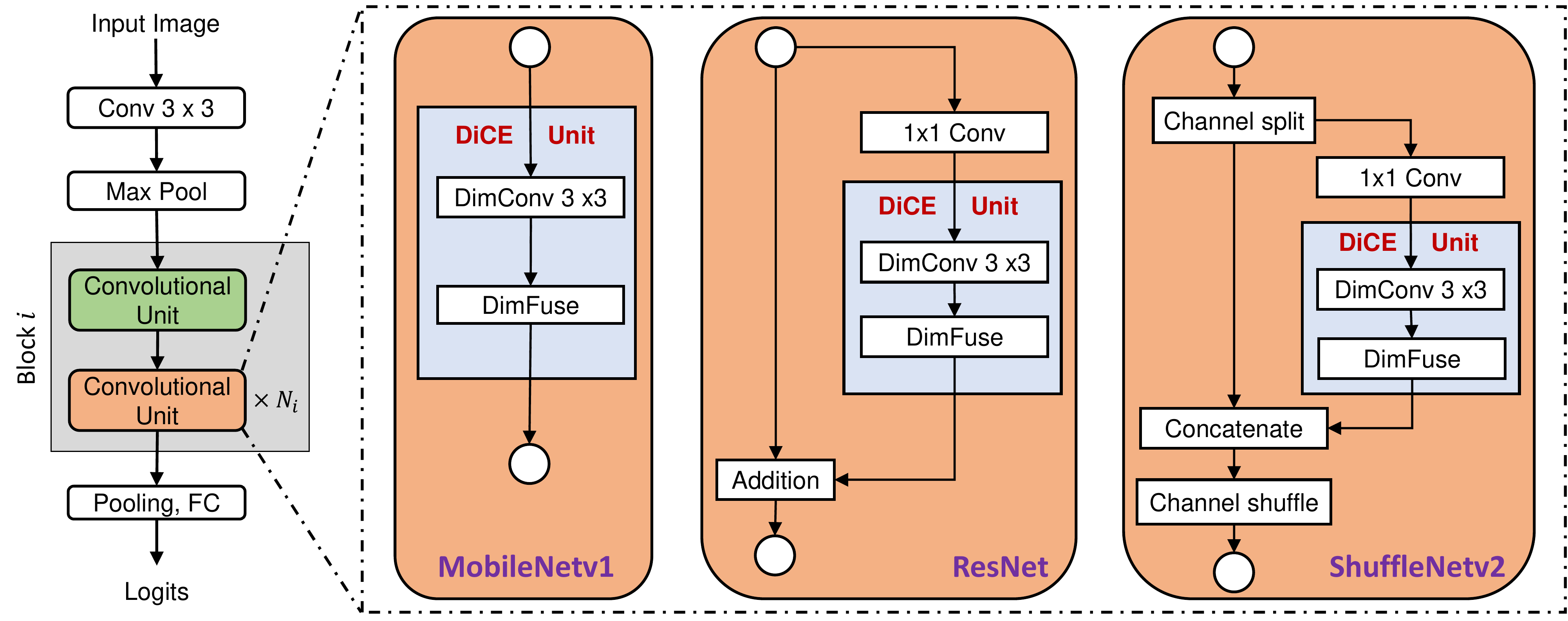}
    \caption{\textbf{\dice~unit in different architecture designs} for the task of image classification on the ImageNet dataset. 
    Green and boxes are with and without stride, respectively. Here, $N_i=\{3, 7, 3\}$ for $i=\{1, 2, 3\}$. See Appendix \ref{sec:append_arch} for detailed architecture specification at different complexity levels.}
    \label{fig:archDiagram}
\end{figure*}
\begin{figure}[t!]
    \centering
    \includegraphics[width=0.9\columnwidth]{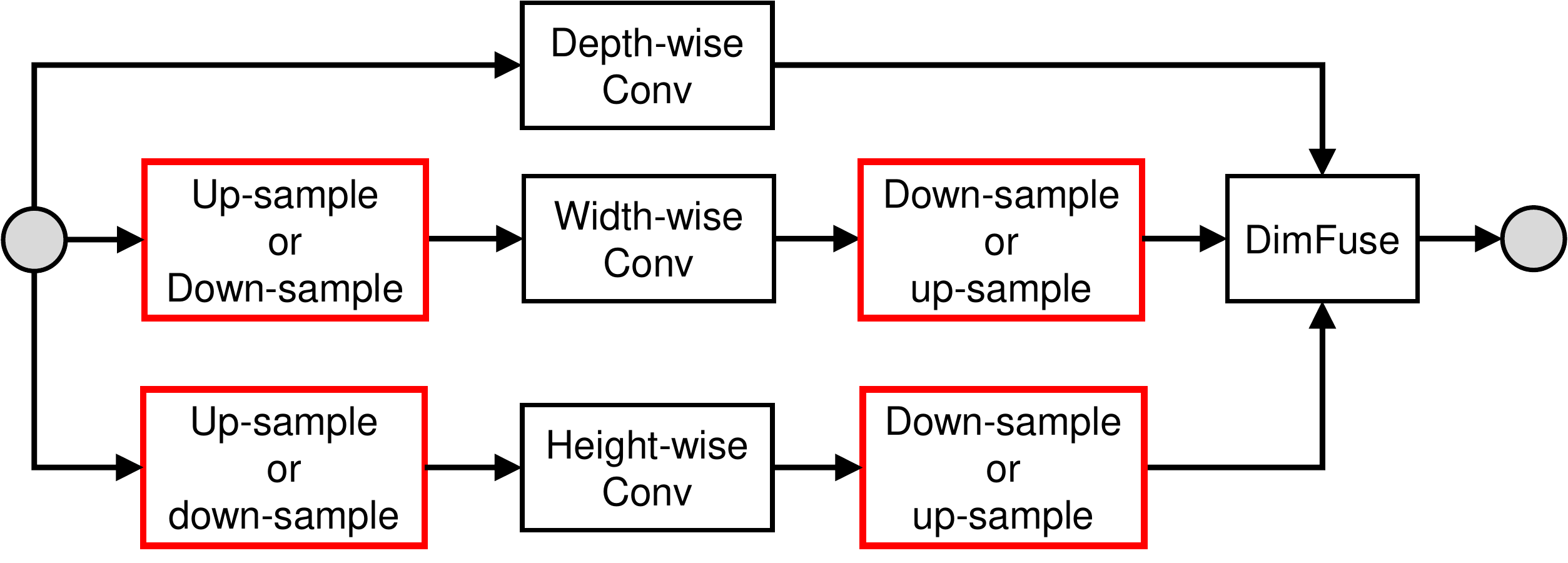}
    \caption{\dice~unit for arbitrary sized input.}
    \label{fig:diceBlock}
\end{figure}

\subsection{\dice~Unit for Arbitrary Sized Inputs}
The \dice~unit stacks \dimconv~and \ecf~to encode spatial and channel-wise information in the input tensor efficiently. However, the two kernels (i.e., $\mathbf{k}_H$ and $\mathbf{k}_W$) in \dimconv~unit correspond to spatial dimensions of the input tensor. This may pose a challenge when spatial dimensions of the input tensor are different from the ones with which the network is trained. To make \dice~units invariant to spatial dimensions of the input tensor, we dynamically scale (either up-sample or down-sample) the height or width dimension of the input tensor to the height or width of the input tensor used in the pretrained network. The resultant tensors are then scaled (either down-sampled or up-sampled) back to their original size before being fed to \ecf; this makes the \dice~unit invariant to input tensor size. Figure \ref{fig:diceBlock} sketches the \dice~unit with dynamic scaling. Results in Section \ref{sec:results}, especially object detection and semantic segmentation, show that the \dice~unit can handle arbitrary sized inputs.

\subsection{\arch~Architecture}
\label{ssec:archDetails}
\dice~units are generic and can be easily integrated in any existing network. Figure \ref{fig:archDiagram} visualizes the \dice~unit with different architectures: (1) \textbf{MobileNet} \cite{howard2017mobilenets} stacks separable convolutions (depth-wise convolution followed by point-wise convolution) to learn representations. (2) \textbf{ResNet} \cite{he2016deep} introduces the bottleneck unit with residual connections to train very deep networks. The bottleneck unit is a stack of three convolutional layers: one $3\times 3$ depth-wise convolutional layer\footnote{Our focus is on efficient network. Therefore, we replace standard convolutional layer with depth-wise convolutional layer.} surrounded by two point-wise convolutions. This block can be viewed as a point-wise convolution followed by separable convolution. (3) \textbf{ShuffleNetv2} \cite{ma2018shufflenet} is a state-of-the-art efficient network that outperforms other efficient networks, including MobileNetv2 \cite{sandler2018mobilenetv2}. ShuffleNetv2's unit stacks a point-wise convolution and separable convolution. It also uses channel split and shuffle to promote feature reuse. 

To illustrate the performance benefits and generic nature of the \dice~unit over separable convolutions, we replace separable convolutions with the \dice~unit in these architectures. Our empirical results in Section \ref{sec:genEff} shows that the \dice~unit with ShuffleNetv2's architecture delivers the best performance. Therefore, we choose the ShuffleNetv2 \cite{ma2018shufflenet} architecture and call the resultant network \arch~(ShuffleNetv2 with the \dice~unit).

\vspace{1mm}
\noindent \textbf{CUDA Implementation:} \dimconv~applies $D$, $W$, and $H$ depth-wise, width-wise and height-wise convolutional kernels to the input tensor $\mathbf{X}$ to aggregate information from different dimensions of the tensor, respectively. A standard solution would be to apply each kernel independently to the tensor and then concatenate their results, as shown in Figure \ref{fig:unoptimzied_kernel}. Another solution would be to apply all kernels simultaneously, as shown in Figure \ref{fig:optimzied_kernel}. Compared to three CUDA kernel calls in the former solution, the later one requires one kernel call, thus reducing the kernel launch time. Also, each thread in the CUDA kernel process $3n^2$ elements compared to $n^2$ elements in the former solution for $n\times n$ convolutional kernels. This maximizes the work done per thread and improves speed. Our results in Section \ref{sec:resultsImg} shows that \arch~is accurate and fast compared to state-of-the-art methods, including neural search-based methods. 

\begin{figure*}
    \centering
    \begin{subfigure}[b]{1.49\columnwidth}
        \begin{subfigure}[b]{0.32\columnwidth}
            \resizebox{!}{100px}{
                \tdplotsetmaincoords{70}{110}

\newcommand{\optimized}{
    \begin{tikzpicture}[tdplot_main_coords]
        \draw (-3.5,-2,-1.5) -- (1, -2,-1.5) -- (1, 1, -1.5) -- (-3.5, 1, -1.5) -- cycle;
        \draw (-3.5,-2, 0.75) -- (1,-2,0.75) -- (1, 1, 0.75) -- (-3.5, 1, 0.75) -- cycle;
        \draw[-latex]  (-3.5,-2,-1.5) -- (-3.5,-2, 1.5) node[left] {\bfseries \large Height};
        \draw (1,-2, 0.75) -- (1,-2,-1.5);
        \draw (1, 1, 0.75) -- (1, 1,-1.5);
        \draw (-3.5, 1, 0.75) -- (-3.5, 1, -1.5);
        \draw[-latex] (1, -2,-1.5) -- (1, 2, -1.5) node[below] {\bfseries \large Depth};
        \draw[-latex] (-3.5,-2,-1.5) -- (2, -2,-1.5) node[below] {\bfseries \large Width};
        \draw[fill=heightwise,opacity=0.3] (0,-1, -1) -- (0,-1,1) -- (0,1,1) -- (0,1,-1) -- cycle; 
        \draw[fill=widthwise,opacity=0.4] (-1,-1,0) -- (-1,1,0) -- (1,1,0) -- (1,-1,0) -- cycle; 
        \draw[fill=depthwise,opacity=0.5] (-1,0,-1) -- (-1,0,1) -- (1,0,1) -- (1,0,-1) -- cycle; 
        \draw[red, fill=red] (0,0) circle (2pt);
    \end{tikzpicture}
}

\newcommand{\dimDepth}{
    \begin{tikzpicture}[tdplot_main_coords]
        \draw (-3.5,-2,-1.5) -- (1, -2,-1.5) -- (1, 1, -1.5) -- (-3.5, 1, -1.5) -- cycle;
        \draw (-3.5,-2, 0.75) -- (1,-2,0.75) -- (1, 1, 0.75) -- (-3.5, 1, 0.75) -- cycle;
        \draw[-latex]  (-3.5,-2,-1.5) -- (-3.5,-2, 1.5) node[left] {\bfseries \large Height};
        \draw (1,-2, 0.75) -- (1,-2,-1.5);
        \draw (1, 1, 0.75) -- (1, 1,-1.5);
        \draw (-3.5, 1, 0.75) -- (-3.5, 1, -1.5);
        \draw[-latex] (1, -2,-1.5) -- (1, 2, -1.5) node[below] {\bfseries \large Depth};
        \draw[-latex] (-3.5,-2,-1.5) -- (2, -2,-1.5) node[below] {\bfseries \large Width};
        \draw[fill=depthwise,opacity=0.5] (-1,0,-1) -- (-1,0,1) -- (1,0,1) -- (1,0,-1) -- cycle; 
        \draw[red, fill=red] (0,0) circle (2pt);
    \end{tikzpicture}
}

\newcommand{\dimWidth}{
    \begin{tikzpicture}[tdplot_main_coords]
        \draw (-3.5,-2,-1.5) -- (1, -2,-1.5) -- (1, 1, -1.5) -- (-3.5, 1, -1.5) -- cycle;
        \draw (-3.5,-2, 0.75) -- (1,-2,0.75) -- (1, 1, 0.75) -- (-3.5, 1, 0.75) -- cycle;
        \draw[-latex]  (-3.5,-2,-1.5) -- (-3.5,-2, 1.5) node[left] {\bfseries \large Height};
        \draw (1,-2, 0.75) -- (1,-2,-1.5);
        \draw (1, 1, 0.75) -- (1, 1,-1.5);
        \draw (-3.5, 1, 0.75) -- (-3.5, 1, -1.5);
        \draw[-latex] (1, -2,-1.5) -- (1, 2, -1.5) node[below] {\bfseries \large Depth};
        \draw[-latex] (-3.5,-2,-1.5) -- (2, -2,-1.5) node[below] {\bfseries \large Width};
        \draw[fill=widthwise,opacity=0.4] (-1,-1,0) -- (-1,1,0) -- (1,1,0) -- (1,-1,0) -- cycle;
        \draw[red, fill=red] (0,0) circle (2pt);
    \end{tikzpicture}
}

\newcommand{\dimHeight}{
    \begin{tikzpicture}[tdplot_main_coords]
        \draw (-3.5,-2,-1.5) -- (1, -2,-1.5) -- (1, 1, -1.5) -- (-3.5, 1, -1.5) -- cycle;
        \draw (-3.5,-2, 0.75) -- (1,-2,0.75) -- (1, 1, 0.75) -- (-3.5, 1, 0.75) -- cycle;
        \draw[-latex]  (-3.5,-2,-1.5) -- (-3.5,-2, 1.5) node[left] {\bfseries \large Height};
        \draw (1,-2, 0.75) -- (1,-2,-1.5);
        \draw (1, 1, 0.75) -- (1, 1,-1.5);
        \draw (-3.5, 1, 0.75) -- (-3.5, 1, -1.5);
        \draw[-latex] (1, -2,-1.5) -- (1, 2, -1.5) node[below] {\bfseries \large Depth};
        \draw[-latex] (-3.5,-2,-1.5) -- (2, -2,-1.5) node[below] {\bfseries \large Width};
        \draw[fill=heightwise,opacity=0.3] (0,-1, -1) -- (0,-1,1) -- (0,1,1) -- (0,1,-1) -- cycle; 
        \draw[red, fill=red] (0,0) circle (2pt);
    \end{tikzpicture}
}\dimDepth
            }
        \end{subfigure}
        \hfill
        \begin{subfigure}[b]{0.32\columnwidth}
            \resizebox{!}{100px}{
                \tdplotsetmaincoords{70}{110}

\newcommand{\optimized}{
    \begin{tikzpicture}[tdplot_main_coords]
        \draw (-3.5,-2,-1.5) -- (1, -2,-1.5) -- (1, 1, -1.5) -- (-3.5, 1, -1.5) -- cycle;
        \draw (-3.5,-2, 0.75) -- (1,-2,0.75) -- (1, 1, 0.75) -- (-3.5, 1, 0.75) -- cycle;
        \draw[-latex]  (-3.5,-2,-1.5) -- (-3.5,-2, 1.5) node[left] {\bfseries \large Height};
        \draw (1,-2, 0.75) -- (1,-2,-1.5);
        \draw (1, 1, 0.75) -- (1, 1,-1.5);
        \draw (-3.5, 1, 0.75) -- (-3.5, 1, -1.5);
        \draw[-latex] (1, -2,-1.5) -- (1, 2, -1.5) node[below] {\bfseries \large Depth};
        \draw[-latex] (-3.5,-2,-1.5) -- (2, -2,-1.5) node[below] {\bfseries \large Width};
        \draw[fill=heightwise,opacity=0.3] (0,-1, -1) -- (0,-1,1) -- (0,1,1) -- (0,1,-1) -- cycle; 
        \draw[fill=widthwise,opacity=0.4] (-1,-1,0) -- (-1,1,0) -- (1,1,0) -- (1,-1,0) -- cycle; 
        \draw[fill=depthwise,opacity=0.5] (-1,0,-1) -- (-1,0,1) -- (1,0,1) -- (1,0,-1) -- cycle; 
        \draw[red, fill=red] (0,0) circle (2pt);
    \end{tikzpicture}
}

\newcommand{\dimDepth}{
    \begin{tikzpicture}[tdplot_main_coords]
        \draw (-3.5,-2,-1.5) -- (1, -2,-1.5) -- (1, 1, -1.5) -- (-3.5, 1, -1.5) -- cycle;
        \draw (-3.5,-2, 0.75) -- (1,-2,0.75) -- (1, 1, 0.75) -- (-3.5, 1, 0.75) -- cycle;
        \draw[-latex]  (-3.5,-2,-1.5) -- (-3.5,-2, 1.5) node[left] {\bfseries \large Height};
        \draw (1,-2, 0.75) -- (1,-2,-1.5);
        \draw (1, 1, 0.75) -- (1, 1,-1.5);
        \draw (-3.5, 1, 0.75) -- (-3.5, 1, -1.5);
        \draw[-latex] (1, -2,-1.5) -- (1, 2, -1.5) node[below] {\bfseries \large Depth};
        \draw[-latex] (-3.5,-2,-1.5) -- (2, -2,-1.5) node[below] {\bfseries \large Width};
        \draw[fill=depthwise,opacity=0.5] (-1,0,-1) -- (-1,0,1) -- (1,0,1) -- (1,0,-1) -- cycle; 
        \draw[red, fill=red] (0,0) circle (2pt);
    \end{tikzpicture}
}

\newcommand{\dimWidth}{
    \begin{tikzpicture}[tdplot_main_coords]
        \draw (-3.5,-2,-1.5) -- (1, -2,-1.5) -- (1, 1, -1.5) -- (-3.5, 1, -1.5) -- cycle;
        \draw (-3.5,-2, 0.75) -- (1,-2,0.75) -- (1, 1, 0.75) -- (-3.5, 1, 0.75) -- cycle;
        \draw[-latex]  (-3.5,-2,-1.5) -- (-3.5,-2, 1.5) node[left] {\bfseries \large Height};
        \draw (1,-2, 0.75) -- (1,-2,-1.5);
        \draw (1, 1, 0.75) -- (1, 1,-1.5);
        \draw (-3.5, 1, 0.75) -- (-3.5, 1, -1.5);
        \draw[-latex] (1, -2,-1.5) -- (1, 2, -1.5) node[below] {\bfseries \large Depth};
        \draw[-latex] (-3.5,-2,-1.5) -- (2, -2,-1.5) node[below] {\bfseries \large Width};
        \draw[fill=widthwise,opacity=0.4] (-1,-1,0) -- (-1,1,0) -- (1,1,0) -- (1,-1,0) -- cycle;
        \draw[red, fill=red] (0,0) circle (2pt);
    \end{tikzpicture}
}

\newcommand{\dimHeight}{
    \begin{tikzpicture}[tdplot_main_coords]
        \draw (-3.5,-2,-1.5) -- (1, -2,-1.5) -- (1, 1, -1.5) -- (-3.5, 1, -1.5) -- cycle;
        \draw (-3.5,-2, 0.75) -- (1,-2,0.75) -- (1, 1, 0.75) -- (-3.5, 1, 0.75) -- cycle;
        \draw[-latex]  (-3.5,-2,-1.5) -- (-3.5,-2, 1.5) node[left] {\bfseries \large Height};
        \draw (1,-2, 0.75) -- (1,-2,-1.5);
        \draw (1, 1, 0.75) -- (1, 1,-1.5);
        \draw (-3.5, 1, 0.75) -- (-3.5, 1, -1.5);
        \draw[-latex] (1, -2,-1.5) -- (1, 2, -1.5) node[below] {\bfseries \large Depth};
        \draw[-latex] (-3.5,-2,-1.5) -- (2, -2,-1.5) node[below] {\bfseries \large Width};
        \draw[fill=heightwise,opacity=0.3] (0,-1, -1) -- (0,-1,1) -- (0,1,1) -- (0,1,-1) -- cycle; 
        \draw[red, fill=red] (0,0) circle (2pt);
    \end{tikzpicture}
}\dimWidth
            }
        \end{subfigure}
        \hfill
        \begin{subfigure}[b]{0.32\columnwidth}
            \resizebox{!}{100px}{
                \tdplotsetmaincoords{70}{110}

\newcommand{\optimized}{
    \begin{tikzpicture}[tdplot_main_coords]
        \draw (-3.5,-2,-1.5) -- (1, -2,-1.5) -- (1, 1, -1.5) -- (-3.5, 1, -1.5) -- cycle;
        \draw (-3.5,-2, 0.75) -- (1,-2,0.75) -- (1, 1, 0.75) -- (-3.5, 1, 0.75) -- cycle;
        \draw[-latex]  (-3.5,-2,-1.5) -- (-3.5,-2, 1.5) node[left] {\bfseries \large Height};
        \draw (1,-2, 0.75) -- (1,-2,-1.5);
        \draw (1, 1, 0.75) -- (1, 1,-1.5);
        \draw (-3.5, 1, 0.75) -- (-3.5, 1, -1.5);
        \draw[-latex] (1, -2,-1.5) -- (1, 2, -1.5) node[below] {\bfseries \large Depth};
        \draw[-latex] (-3.5,-2,-1.5) -- (2, -2,-1.5) node[below] {\bfseries \large Width};
        \draw[fill=heightwise,opacity=0.3] (0,-1, -1) -- (0,-1,1) -- (0,1,1) -- (0,1,-1) -- cycle; 
        \draw[fill=widthwise,opacity=0.4] (-1,-1,0) -- (-1,1,0) -- (1,1,0) -- (1,-1,0) -- cycle; 
        \draw[fill=depthwise,opacity=0.5] (-1,0,-1) -- (-1,0,1) -- (1,0,1) -- (1,0,-1) -- cycle; 
        \draw[red, fill=red] (0,0) circle (2pt);
    \end{tikzpicture}
}

\newcommand{\dimDepth}{
    \begin{tikzpicture}[tdplot_main_coords]
        \draw (-3.5,-2,-1.5) -- (1, -2,-1.5) -- (1, 1, -1.5) -- (-3.5, 1, -1.5) -- cycle;
        \draw (-3.5,-2, 0.75) -- (1,-2,0.75) -- (1, 1, 0.75) -- (-3.5, 1, 0.75) -- cycle;
        \draw[-latex]  (-3.5,-2,-1.5) -- (-3.5,-2, 1.5) node[left] {\bfseries \large Height};
        \draw (1,-2, 0.75) -- (1,-2,-1.5);
        \draw (1, 1, 0.75) -- (1, 1,-1.5);
        \draw (-3.5, 1, 0.75) -- (-3.5, 1, -1.5);
        \draw[-latex] (1, -2,-1.5) -- (1, 2, -1.5) node[below] {\bfseries \large Depth};
        \draw[-latex] (-3.5,-2,-1.5) -- (2, -2,-1.5) node[below] {\bfseries \large Width};
        \draw[fill=depthwise,opacity=0.5] (-1,0,-1) -- (-1,0,1) -- (1,0,1) -- (1,0,-1) -- cycle; 
        \draw[red, fill=red] (0,0) circle (2pt);
    \end{tikzpicture}
}

\newcommand{\dimWidth}{
    \begin{tikzpicture}[tdplot_main_coords]
        \draw (-3.5,-2,-1.5) -- (1, -2,-1.5) -- (1, 1, -1.5) -- (-3.5, 1, -1.5) -- cycle;
        \draw (-3.5,-2, 0.75) -- (1,-2,0.75) -- (1, 1, 0.75) -- (-3.5, 1, 0.75) -- cycle;
        \draw[-latex]  (-3.5,-2,-1.5) -- (-3.5,-2, 1.5) node[left] {\bfseries \large Height};
        \draw (1,-2, 0.75) -- (1,-2,-1.5);
        \draw (1, 1, 0.75) -- (1, 1,-1.5);
        \draw (-3.5, 1, 0.75) -- (-3.5, 1, -1.5);
        \draw[-latex] (1, -2,-1.5) -- (1, 2, -1.5) node[below] {\bfseries \large Depth};
        \draw[-latex] (-3.5,-2,-1.5) -- (2, -2,-1.5) node[below] {\bfseries \large Width};
        \draw[fill=widthwise,opacity=0.4] (-1,-1,0) -- (-1,1,0) -- (1,1,0) -- (1,-1,0) -- cycle;
        \draw[red, fill=red] (0,0) circle (2pt);
    \end{tikzpicture}
}

\newcommand{\dimHeight}{
    \begin{tikzpicture}[tdplot_main_coords]
        \draw (-3.5,-2,-1.5) -- (1, -2,-1.5) -- (1, 1, -1.5) -- (-3.5, 1, -1.5) -- cycle;
        \draw (-3.5,-2, 0.75) -- (1,-2,0.75) -- (1, 1, 0.75) -- (-3.5, 1, 0.75) -- cycle;
        \draw[-latex]  (-3.5,-2,-1.5) -- (-3.5,-2, 1.5) node[left] {\bfseries \large Height};
        \draw (1,-2, 0.75) -- (1,-2,-1.5);
        \draw (1, 1, 0.75) -- (1, 1,-1.5);
        \draw (-3.5, 1, 0.75) -- (-3.5, 1, -1.5);
        \draw[-latex] (1, -2,-1.5) -- (1, 2, -1.5) node[below] {\bfseries \large Depth};
        \draw[-latex] (-3.5,-2,-1.5) -- (2, -2,-1.5) node[below] {\bfseries \large Width};
        \draw[fill=heightwise,opacity=0.3] (0,-1, -1) -- (0,-1,1) -- (0,1,1) -- (0,1,-1) -- cycle; 
        \draw[red, fill=red] (0,0) circle (2pt);
    \end{tikzpicture}
}\dimHeight
            }
        \end{subfigure}
        \caption{Unoptimized (Left to right: depth-wise, height-wise, and width-wise)}
        \label{fig:unoptimzied_kernel}
    \end{subfigure}
    \hfill
    \begin{subfigure}[b]{0.49\columnwidth}
        \resizebox{!}{100px}{
            \tdplotsetmaincoords{70}{110}

\newcommand{\optimized}{
    \begin{tikzpicture}[tdplot_main_coords]
        \draw (-3.5,-2,-1.5) -- (1, -2,-1.5) -- (1, 1, -1.5) -- (-3.5, 1, -1.5) -- cycle;
        \draw (-3.5,-2, 0.75) -- (1,-2,0.75) -- (1, 1, 0.75) -- (-3.5, 1, 0.75) -- cycle;
        \draw[-latex]  (-3.5,-2,-1.5) -- (-3.5,-2, 1.5) node[left] {\bfseries \large Height};
        \draw (1,-2, 0.75) -- (1,-2,-1.5);
        \draw (1, 1, 0.75) -- (1, 1,-1.5);
        \draw (-3.5, 1, 0.75) -- (-3.5, 1, -1.5);
        \draw[-latex] (1, -2,-1.5) -- (1, 2, -1.5) node[below] {\bfseries \large Depth};
        \draw[-latex] (-3.5,-2,-1.5) -- (2, -2,-1.5) node[below] {\bfseries \large Width};
        \draw[fill=heightwise,opacity=0.3] (0,-1, -1) -- (0,-1,1) -- (0,1,1) -- (0,1,-1) -- cycle; 
        \draw[fill=widthwise,opacity=0.4] (-1,-1,0) -- (-1,1,0) -- (1,1,0) -- (1,-1,0) -- cycle; 
        \draw[fill=depthwise,opacity=0.5] (-1,0,-1) -- (-1,0,1) -- (1,0,1) -- (1,0,-1) -- cycle; 
        \draw[red, fill=red] (0,0) circle (2pt);
    \end{tikzpicture}
}

\newcommand{\dimDepth}{
    \begin{tikzpicture}[tdplot_main_coords]
        \draw (-3.5,-2,-1.5) -- (1, -2,-1.5) -- (1, 1, -1.5) -- (-3.5, 1, -1.5) -- cycle;
        \draw (-3.5,-2, 0.75) -- (1,-2,0.75) -- (1, 1, 0.75) -- (-3.5, 1, 0.75) -- cycle;
        \draw[-latex]  (-3.5,-2,-1.5) -- (-3.5,-2, 1.5) node[left] {\bfseries \large Height};
        \draw (1,-2, 0.75) -- (1,-2,-1.5);
        \draw (1, 1, 0.75) -- (1, 1,-1.5);
        \draw (-3.5, 1, 0.75) -- (-3.5, 1, -1.5);
        \draw[-latex] (1, -2,-1.5) -- (1, 2, -1.5) node[below] {\bfseries \large Depth};
        \draw[-latex] (-3.5,-2,-1.5) -- (2, -2,-1.5) node[below] {\bfseries \large Width};
        \draw[fill=depthwise,opacity=0.5] (-1,0,-1) -- (-1,0,1) -- (1,0,1) -- (1,0,-1) -- cycle; 
        \draw[red, fill=red] (0,0) circle (2pt);
    \end{tikzpicture}
}

\newcommand{\dimWidth}{
    \begin{tikzpicture}[tdplot_main_coords]
        \draw (-3.5,-2,-1.5) -- (1, -2,-1.5) -- (1, 1, -1.5) -- (-3.5, 1, -1.5) -- cycle;
        \draw (-3.5,-2, 0.75) -- (1,-2,0.75) -- (1, 1, 0.75) -- (-3.5, 1, 0.75) -- cycle;
        \draw[-latex]  (-3.5,-2,-1.5) -- (-3.5,-2, 1.5) node[left] {\bfseries \large Height};
        \draw (1,-2, 0.75) -- (1,-2,-1.5);
        \draw (1, 1, 0.75) -- (1, 1,-1.5);
        \draw (-3.5, 1, 0.75) -- (-3.5, 1, -1.5);
        \draw[-latex] (1, -2,-1.5) -- (1, 2, -1.5) node[below] {\bfseries \large Depth};
        \draw[-latex] (-3.5,-2,-1.5) -- (2, -2,-1.5) node[below] {\bfseries \large Width};
        \draw[fill=widthwise,opacity=0.4] (-1,-1,0) -- (-1,1,0) -- (1,1,0) -- (1,-1,0) -- cycle;
        \draw[red, fill=red] (0,0) circle (2pt);
    \end{tikzpicture}
}

\newcommand{\dimHeight}{
    \begin{tikzpicture}[tdplot_main_coords]
        \draw (-3.5,-2,-1.5) -- (1, -2,-1.5) -- (1, 1, -1.5) -- (-3.5, 1, -1.5) -- cycle;
        \draw (-3.5,-2, 0.75) -- (1,-2,0.75) -- (1, 1, 0.75) -- (-3.5, 1, 0.75) -- cycle;
        \draw[-latex]  (-3.5,-2,-1.5) -- (-3.5,-2, 1.5) node[left] {\bfseries \large Height};
        \draw (1,-2, 0.75) -- (1,-2,-1.5);
        \draw (1, 1, 0.75) -- (1, 1,-1.5);
        \draw (-3.5, 1, 0.75) -- (-3.5, 1, -1.5);
        \draw[-latex] (1, -2,-1.5) -- (1, 2, -1.5) node[below] {\bfseries \large Depth};
        \draw[-latex] (-3.5,-2,-1.5) -- (2, -2,-1.5) node[below] {\bfseries \large Width};
        \draw[fill=heightwise,opacity=0.3] (0,-1, -1) -- (0,-1,1) -- (0,1,1) -- (0,1,-1) -- cycle; 
        \draw[red, fill=red] (0,0) circle (2pt);
    \end{tikzpicture}
}\optimized
        }
        \caption{Optimized}
        \label{fig:optimzied_kernel}
    \end{subfigure}
    \caption{\textbf{Implementation of dimension-wise convolution (\dimconv)}. In (a), each kernel is applied to a pixel (represented by \textcolor{red}{red} dot) independently. In (b), all kernels are applied to a pixel simultaneously, allowing us to aggregate the information from tensor efficiently. Convolutional kernels are highlighted in color (\textbf{\textcolor{depthwise}{depth-}}, \textbf{\textcolor{widthwise}{width-}}, and \textbf{\textcolor{heightwise}{height-}}wise).}
    \label{fig:cuda_implement}
\end{figure*}

\section{Experimental Set-up}
Following most architecture designs (e.g. \cite{he2016deep, howard2017mobilenets, mehta2018espnetv2}), we evaluate the generic nature of the \dice~unit on the ImageNet dataset in Section \ref{sec:genEff}. We integrate the \dice~unit in different image classification architectures (Figure \ref{fig:archDiagram}) and study the impact on efficiency and accuracy. We also study the importance of the two main components of the \dice~unit, i.e. \dimconv~and \ecf, and show that \dice~units are more effective than separable convolutions \cite{howard2017mobilenets}. In Section \ref{sec:resultsImg}, we evaluate the image classification performance of \arch~on the ImageNet dataset and show that \arch~delivers similar or better performance than state-of-the-art efficient networks, including neural search-based methods. In Section \ref{sec:results}, we evaluate task-level generalization ability of \arch~on three different visual recognition tasks, i.e. object detection, semantic segmentation, and multi-object classification, that are often used in resource-constrained devices. We demonstrate that \arch~generalizes better than existing efficient networks that are built using separable convolutions. 

\vspace{1mm}
\noindent \textbf{Datasets:} We use following datasets in our experiments.

\vspace{1mm}
\noindent \textit{\bfseries Image classification:} For \emph{single label} image classification, we use ImageNet-1K classification dataset \cite{ILSVRC15}. This dataset consists of 1.28M training and 50K validation images. All networks on this dataset are trained from scratch. For \emph{multi-label} classification, we use MS-COCO dataset \cite{lin2014microsoft} that has 2.9 labels (on an average) per image. We use the same training and validation splits as in \cite{mehta2018espnetv2}.

\vspace{1mm}
\noindent \textit{\bfseries Object detection:} We use MS-COCO \cite{lin2014microsoft} and PASCAL VOC 2007 \cite{pascalvoc2007} datasets for evaluating on the task of object detection. Following a standard convention for training on PASCAL VOC 2007 dataset, we augment it with PASCAL VOC 2012 \cite{pascal2012} and PASCAL VOC 2007 \textit{trainval} set for training and evaluate the performance on PASCAL VOC 2007 \textit{test} set. 

\vspace{1mm}
\noindent \textit{Semantic segmentation:} We use PASCAL VOC 2012 \cite{pascal2012} dataset for this task. Following a standard convention, we use additional images for training from \cite{hariharan2011semantic} and \cite{lin2014microsoft}. Similar to MobileNetv2 \cite{sandler2018mobilenetv2}, we evaluate the performance on the validation set.

\vspace{1mm}
\noindent \textbf{Efficiency metric:} We measure efficiency in terms of the number of floating point operations (FLOPs) and inference time. We use PyTorch for training our networks.

\section{Evaluating \dice~Unit on ImageNet}
\label{sec:genEff}
We first evaluate the two important properties of the \dice~unit, i.e. generic and efficiency, in Section \ref{ssec:diceGeneric}. To evaluate this, we replace separable convolutions \cite{howard2017mobilenets} in different architectures with the \dice~unit (Figure \ref{fig:archDiagram}). We then study the importance of each component of the \dice~unit, \dimconv~and \ecf, in Section \ref{ssec:diceEff}. Recent studies (e.g., MobileNetv3 \cite{howard2019searching} and MNASNet \cite{tan2018mnasnet}) uses several different methods, such as exponential moving average (EMA) and large batch sizes, to improve the performance. In Section \ref{ssec:opt_hyper}, we study the effect of these methods on the performance of \arch. 

In these experiment, we use the ImageNet dataset \cite{ILSVRC15}. In Section \ref{ssec:diceGeneric} and Section \ref{ssec:diceEff}, we follow the experimental setup of ESPNetv2 \cite{mehta2018espnetv2} and ShuffleNetv2 \cite{ma2018shufflenet} (fewer training epochs with smaller batch size) while in Section \ref{ssec:opt_hyper}, we follow experimental set-up similar to MobileNets \cite{sandler2018mobilenetv2,howard2019searching} (longer training with larger batch size).

\subsection{\dice~Unit vs. Separable Convolutions}
\label{ssec:diceGeneric}
Table \ref{tab:diffArchs} shows the performance of the \dice~unit with different architectures at different FLOP ranges. When separable convolutions are replaced with the \dice~unit in MobileNet architecture, we observe significant gains in performance both in terms of accuracy and efficiency. Since this architecture does not employ any advanced methods (e.g., residual connections and channel shuffle) to improve performance, it allows us to understand the ``true" gains of the \dice~unit over separable convolutions. 

When we replace separable convolutions with the \dice~unit in ResNet and ShuffleNetv2, we observe significant improvements especially for small-(25-60 MFLOPs) and medium-sized (120-170 MFLOPs) models. For instance, the \dice~unit improved the performance of ShuffleNetv2 by about 3\% for small-sized model (about 40 MFLOPs). Similarly, ShuffleNetv2 with the \dice~unit requires 24 million fewer FLOPs to achieve the same accuracy as with separable convolutions for medium-sized model (120-170 MFLOPs). These results suggests that the \dice~unit is generic and learns better representations than separable convolutions.

\begin{table}[t!]
    \centering
    \resizebox{\columnwidth}{!}{
    \begin{tabular}{lrrcrrcrr}
    \toprule
     \multicolumn{1}{c}{\textbf{FLOP}} & \multicolumn{2}{c}{\textbf{Separable conv}} & & \multicolumn{2}{c}{\dice~\textbf{unit}} & & \multicolumn{2}{c}{\bf Absolute difference}\\
     \multicolumn{1}{c}{\textbf{Range}} & \multicolumn{2}{c}{(\textbf{SC})} & & \multicolumn{2}{c}{(\textbf{DU})} & & \multicolumn{2}{c}{\bf (DU - SC)}\\
     \cmidrule{2-3}\cmidrule{5-6}\cmidrule{8-9}
        \textbf{(in millions)} & Top-1 & FLOPs & & Top-1 & FLOPs & &  Top-1 & FLOPs\\
    \bottomrule
    \midrule
    \multicolumn{9}{c}{\textbf{\textcolor{highlightArch}{MobileNet}} \cite{howard2017mobilenets}} \\
    \midrule
        25-60  & 49.80 & 41 M & & \textbf{52.55} & \textbf{29} M & & \incAcc{2.75} & \redF{12 M} \\
        120-170  & 65.30 & \textbf{162} M & & \textbf{69.05} & 167 M & & \incAcc{3.75} & \incF{5 M} \\
        270-320 & 68.40 & 317 M & & \textbf{70.83} & \textbf{277} M & & \incAcc{2.43} & \redF{40 M} \\
    \bottomrule
    \midrule
    \multicolumn{9}{c}{\textbf{\textcolor{highlightArch}{ResNet}} \cite{he2016deep}} \\
    \midrule
        25-60  & 59.30 & 59 M &  & \textbf{61.35} & \textbf{52} M  & & \incAcc{2.05} & \redF{7 M} \\
        120-170 & 67.80 & 142 M &  & \textbf{67.90} & \textbf{122} M  & & \incAcc{0.10} & \redF{20 M} \\
        270-320  & 70.67 & 302 M &  & \textbf{71.80} & \textbf{300} M  & & \incAcc{1.13} & \redF{2 M} \\
    \bottomrule
    \midrule
    \multicolumn{9}{c}{\textbf{\textcolor{highlightArch}{ShuffleNetv2}} \cite{ma2018shufflenet}} \\
    \midrule
        25-60 & 59.69 & \textbf{41} M &  & \textbf{62.80} & 46 M &  & \incAcc{3.11} & \incF{5 M} \\
        120-170 & 68.14 & 146 M &  & \textbf{68.21} & \textbf{122} M &  & \incAcc{0.07} & \redF{24 M} \\
        270-320 & 71.80 & \textbf{292 M} &  & \textbf{72.90} & 298 M &  & \incAcc{1.10} & \incF{6 M} \\
    \bottomrule
    \end{tabular}
    }
    \caption{Comparison between the \dice~unit and separable convolutions on the ImageNet dataset across \textcolor{highlightArch}{different architectures}.  Models with the \dice~unit requires fewer channels compared to models with separable convolution in order to obtain similar performance. Thus, models with the \dice~unit has fewer FLOPs compared to separable convolutions.}
    \label{tab:diffArchs}
\end{table}

\subsection{Importance of \dimconv~and \ecf}
\label{ssec:diceEff}
To understand the significance of each component of the \dice~unit, we replace \dimconv~with depth-wise convolution and \ecf~with different fusion methods, including point-wise convolutions and squeeze-excitation (SE) unit \cite{hu2018squeeze} and study their combinations for two architectures (ResNet and ShuffleNetv2)\footnote{MobileNet's performance is significantly lower than ResNet and ShuffleNetv2, therefore, we do not use MobileNet for these experiments.}. In these experiments, we study efficient models by restricting the computational budget between 120 and 150 MFLOPs.

\vspace{1mm}
\noindent \textbf{Importance of \dimconv:} We replace depth-wise convolutional layers with \dimconv~in ResNet and ShuffleNetv2 architectures. Table \ref{tab:depthVsDim} shows that these networks with \dimconv~require about 10-11 million fewer FLOPs to achieve similar accuracy as the depth-wise convolution. These results suggest that encoding spatial and channel-wise information independently in \dimconv~helps learning better representations compared to encoding only spatial information in depth-wise convolution.

\vspace{1mm}
\noindent \textbf{Importance of \ecf:} To understand the effect of \ecf, we replace \ecf~with two widely used fusion operations, i.e., point-wise convolution and SE unit. Table \ref{tab:diffFuse} summarizes the results. Compared to the widely used combination of depth-wise and point-wise convolutions (or separable convolution), the combination of \dimconv~and \ecf~(or the \dice~unit) is the most effective and improves the efficiency of networks by 15-20\% with little or no impact on accuracy. 

The combination of depth-wise and \ecf~is not as effective as \dimconv~and \ecf. \dimconv~encodes local spatial and channel-wise information, which enables the \dice~unit to use a less complex fusion method (\ecf) for encoding global information. Unlike \dimconv, depth-wise convolutions only encode local spatial information and require computationally expensive point-wise convolutions to encode global information. 

When point-wise convolutions are replaced with SE unit, the performance of networks with depth-wise and \dimconv~convolutions dropped significantly. This is because SE unit relies on an existing convolutional unit, such as ResNext \cite{xie2017aggregated}, to encode global spatial and channel-wise information. When SE unit is used as a \textit{replacement} for point-wise convolutions, it fails to effectively encode this information; resulting in significant performance drop.

\begin{table}[t!]
    \centering
    \begin{subtable}[h]{\columnwidth}
        \centering
        \resizebox{0.85\columnwidth}{!}{
        \begin{tabular}{lrrrrr}
            \toprule
            & \multicolumn{2}{c}{\textbf{ResNet}} &  & \multicolumn{2}{c}{\textbf{ShuffleNetv2}}\\ 
            \cmidrule{2-3}\cmidrule{5-6}
            \textbf{Layer} & \textbf{FLOPs} & \textbf{Top-1} &  & \textbf{FLOPs} & \textbf{Top-1}\\
            \midrule
              DWise + Point-wise & 142 M & 67.80 & &  146 M & 68.14 \\ 
              \dimconv~+ Point-wise & \textbf{132 M}  & \textbf{68.10} & & \textbf{135 M} & \textbf{68.45} \\
            \bottomrule
        \end{tabular}
        }
        \caption{Importance of \dimconv. DWise denotes depth-wise conv.}
        \label{tab:depthVsDim}
    \end{subtable}
    \vfill
    \begin{subtable}[h]{\columnwidth}
        \centering
        \resizebox{0.95\columnwidth}{!}{
        \begin{tabular}{lrrrrr}
            \toprule
            & \multicolumn{2}{c}{\textbf{ResNet}} &  & \multicolumn{2}{c}{\textbf{ShuffleNetv2}}\\
            \cmidrule{2-3}\cmidrule{5-6}
            \textbf{Layer} & \textbf{FLOPs} & \textbf{Top-1} &  & \textbf{FLOPs} & \textbf{Top-1}\\
            \midrule
              DWise + Point-wise (Separable) & 142 M & 67.80 & & 146 M & 68.14 \\ 
              DWise + SE & 137 M & 63.90 & & 140 M & 64.70 \\ 
              DWise + Point-wise + SE & 142 M & 68.20 & & 146 M & 68.60 \\ 
              DWise + \ecf & 136 M & 65.90 & &  139 M & 66.80 \\
            \midrule
             \dimconv~ + Point-wise & 132 M  & \textbf{68.10} & & 135 M & \textbf{68.45} \\
             \dimconv~ + SE & 134 M & 64.80 & &  138 M & 65.40 \\ 
             \dimconv~ + Point-wise + SE & 132 M & 67.90 & &  135 M & 68.30 \\ 
             \dimconv~ + \ecf~(\dice~unit) & \textbf{122 M} & 67.90 & &  \textbf{122 M} & 68.21 \\  
             \bottomrule
        \end{tabular}
        }
        \caption{Importance of \ecf. DWise denotes depth-wise conv.}
        \label{tab:diffFuse}
    \end{subtable}
    \caption{Evaluating \dice~unit on the ImageNet dataset. Top-1 accuracy is reported on the validation set. Models with \dice~unit requires fewer channels compared to models with separable convolution in order to obtain similar performance. Therefore, models with \dice~unit has fewer FLOPs compared to separable convolutions.} 
\end{table}

\begin{table}[t!]
    \centering
    \resizebox{\columnwidth}{!}{
    \begin{tabular}{llcrrrrr}
        \toprule
        \textbf{Row} & & \textbf{Network} & \multicolumn{2}{c}{\textbf{ResNet}} &  & \multicolumn{2}{c}{\textbf{ShuffleNetv2}}\\ 
        \cmidrule{4-5}\cmidrule{7-8}
        \# & \textbf{Layer} & \textbf{Width} & \textbf{FLOPs} & \textbf{Top-1} &  & \textbf{FLOPs} & \textbf{Top-1}\\
        \midrule
        R1 &  Point-wise + DWise + \ecf & $1\times$ & 136 M & 65.90 & & 139 M & 66.80 \\ 
        R2 & \ecf~ + DWise + \ecf & $1\times$ & 78 M & 60.10 &   & 78 M & 61.80 \\ 
        R3 & \ecf~ + DWise + \ecf & $4\times$ & 141 M & 66.20 &   & 140 M & 66.90\\
        \midrule
        R4 & Point-wise + \dimconv~ + \ecf & $1\times$ & 122 M & 67.90 & &  122 M & 68.21 \\ 
        R5 & \ecf~ + \dimconv~ + \ecf & $1\times$ & \textbf{72 M} & 62.10 &  &  \textbf{72 M} & 63.70 \\ 
        R6 & \ecf~ + \dimconv~ + \ecf & $4\times$ & 129 M & \textbf{68.20} &  & 132 M & \textbf{69.20}  \\  
         \bottomrule
    \end{tabular}
    }
    \caption{Impact of replacing all pointwise convolutions with \ecf. Here, DWise denotes depth-wise convolution.}
    \label{tab:replaceAllPt}
\end{table}

\begin{table*}[t!]
	\centering
	    \resizebox{1.9\columnwidth}{!}{
	    \begin{tabular}{llrrrrrrr}
	        \toprule[1.5pt]
	        \multicolumn{1}{l}{\multirow{2}{*}{\textbf{Network}}} & \multicolumn{1}{l}{\multirow{2}{*}{\textbf{Type}}} & \multicolumn{7}{c}{\textbf{FLOP ranges (in millions)}} \\
	        \cmidrule[1.25pt]{3-9}
	         &  & \multicolumn{1}{c}{$<$ 10 M} & \multicolumn{1}{c}{10-20 M} & \multicolumn{1}{c}{21-60 M} & \multicolumn{1}{c}{61-90 M} & \multicolumn{1}{c}{91-130 M} & \multicolumn{1}{c}{131-170 M} & \multicolumn{1}{c}{200 -320 M} \\
	         \midrule[1pt]
	         \textbf{MobileNet} \cite{howard2017mobilenets} & Manual & & 41.5 (14) & 56.3 (49) & 59.1 (77) & 61.7 (110) & 65.3 (162) & 68.4 (317) \\
	        \textbf{MobileNetv2} \cite{sandler2018mobilenetv2} & Manual & & 45.5 (\textbf{11}) & 61.0 (50) & 63.9 (71) & 66.4 (107) & 68.7 (153) & 71.8 (300) \\
	       \textbf{ESPNetv2} \cite{mehta2018espnetv2} & Manual & & & & 66.1 (86) & 67.9 (124) & & 72.1 (284) \\
	       \textbf{CondenseNet} \cite{huang2017condensenet} & Manual & & & &  &  & & 71.0 (274)  \\
	       %
	       \textbf{ShuffleNetv2} \cite{ma2018shufflenet} & Manual & 39.1 (8.0) & & 59.7 (41) & & & 68.1 (142) & 71.8 (292) \\
	       \midrule
	       \textbf{MNASNet} \cite{tan2018mnasnet} & NAS & & & & 62.4 (76) & 67.3 (103) & & 74.0 (317) \\
	       \textbf{FBNet} \cite{wu2018fbnet} & NAS & & & & 65.3 (72) & 67.0 (92) & & 74.1 (295) \\
	       \textbf{MobileNetv3} \cite{howard2019searching} & NAS  & & &  & 67.4 (66) &  &  & 75.2 (219) \\
	       \textbf{MixNet} \cite{Tan2019MixConvMD} & NAS & & & &  &  &  & \textbf{75.8} (256) \\
	       \midrule 
	       \textbf{\arch-E150-B512 (Ours)} & Manual & 40.6 (6.5) & 46.2 (14) & 62.8 (46) & 66.5 (70) & 67.8 (98) & 69.5 (139) & 72.9 (298) \\
	       \textbf{\arch-E300-B2048 (Ours)} & Manual & \textbf{43.1} (6.5) & \textbf{48.2} (14) & \textbf{65.1} (46) & \textbf{68.5} (70) & \textbf{69.3} (98) & \textbf{72.0} (139) & 75.7 (298) \\
	        \bottomrule[1.5pt]
	    \end{tabular}
	    }
	\caption{\textbf{Results on the ImageNet dataset.} \arch~delivers similar or better performance than state-of-the-art methods, including neural architecture search (NAS)-based methods. Here, each entry is represented as top-1 accuracy and FLOPs within parentheses. \arch-E150-B512 models are trained for 150 epochs with a batch size of 512 (without EMA and label smoothing) while \arch-E300-B2048  models are trained for 300 epochs with an effective batch size of 2048 (with EMA and label smoothing). }
	\label{tab:imgnetRes}
\end{table*}
\begin{table*}[t!]
    \centering
\resizebox{2\columnwidth}{!}{
    \begin{tabular}{lcccccccccc}
    \toprule[1.5pt]
            & \multicolumn{3}{c}{\bfseries Network statistics} && \multicolumn{2}{c}{\bfseries Device: GTX-960 M (Memory = 4GB)} && \multicolumn{3}{c}{\bfseries Device: GTX-1080 Ti (Memory = 11 GB)} \\
    \cmidrule[1.25pt]{2-4} \cmidrule[1.25pt]{6-7} \cmidrule[1.25pt]{9-11}
    \textbf{Model}       & \textbf{\# Params} & \textbf{\# FLOPs} & \textbf{Top-1 Accuracy} && \textbf{Batch size = 1}  & \textbf{Batch size = 32}      && \textbf{Batch size = 1}  & \textbf{Batch size = 32}  & \textbf{Batch size = 64}   \\
    \midrule[1pt]
    \textbf{MobileNetv2} & 3.5 M  & 300 M   & 71.8  && \textbf{5.6 $\pm$ 0.2 ms} & 114 $\pm$ 0.1 ms && \textbf{5.9 $\pm$ 0.1 ms} & 22.8 $\pm$ 0.1 ms & 44.3 $\pm$ 0.8 ms  \\
    \textbf{ShuffleNetv2} & 3.5 M & 300 M & 71.8 && \textbf{5.8 $\pm$ 0.2 ms} & 80.7 $\pm$ 0.6 ms && \textbf{5.8 $\pm$ 0.2 ms} & \textbf{12.9 $\pm$ 0.1 ms} & \textbf{24.1 $\pm$ 0.4 ms} \\
    \textbf{MobileNetv3} & 5.5 M  & 220 M  & 75.2 && 8.5 $\pm$ 0.2 ms  & Out-of-memory                  && 9.0 $\pm$ 0.3 ms & 20.9 $\pm$ 0.1 ms & 40.2 $\pm$ 0.2 ms \\
    \textbf{\arch~(Ours)} & 5.1 M  & 297 M  & \textbf{75.7}  && \textbf{5.9 $\pm$ 0.1 ms} & \textbf{79 $\pm$ 0.1 ms}  && \textbf{5.7 $\pm$ 0.1 ms} & \textbf{12.8 $\pm$ 0.1 ms} & \textbf{24.1 $\pm$ 0.6 ms} \\
    \bottomrule[1.5pt]
    \end{tabular}
}
    \caption{\textbf{Inference speed}. \arch~and ShuffleNetv2 are comparatively faster than MobileNetv2 and MobileNetv3 on both devices (Mobile GPU: GTX-960M and Desktop GPU: GTX-1080 Ti). Inference results are an average over 100 trials for RGB input images of size $224 \times 224$. We used PyTorch with CUDA 10.2 for measuring speed. MobileNetv2 and ShuffleNetv2 implementations are taken from official PyTorch repository while MobileNetv3's implementation is taken from \cite{rossPytorch}. Since efficient implementations of EESP module in ESPNetv2 and mixed depth-wise convolution in MixNet are not available in PyTorch, we do not compare with these works.}
\end{table*}

\vspace{1mm}
\noindent \textbf{\ecf~replacing  all point-wise convolutions:} The first layer in ShuffleNetv2 and ResNet is a point-wise convolution (Figure \ref{fig:archDiagram}; Section \ref{ssec:archDetails}). Table \ref{tab:diffFuse} shows that \ecf~is an effective replacement for point-wise convolutions. A natural question arises if we can replace the first point-wise convolutional layer in these architectures with \ecf. Table \ref{tab:replaceAllPt} shows the effect of replacing point-wise convolutions with \ecf. For a fixed network width, the number of FLOPs are reduced by 45\% when point-wise convolutions are replaced with \ecf, however, the accuracy drops by about 5-7\%. For similar number of FLOPs, networks with \ecf~achieves higher accuracy. However, such networks (R3 and R6)  are about $4\times$ wider than the networks with point-wise convolutions (R1 and R4); this poses memory constraints for resource-constrained devices. Therefore, we only replace separable convolutions with the \dice~unit while keeping the remaining architecture intact.

\begin{figure}[b!]
    \centering
    \includegraphics[width=0.8\columnwidth]{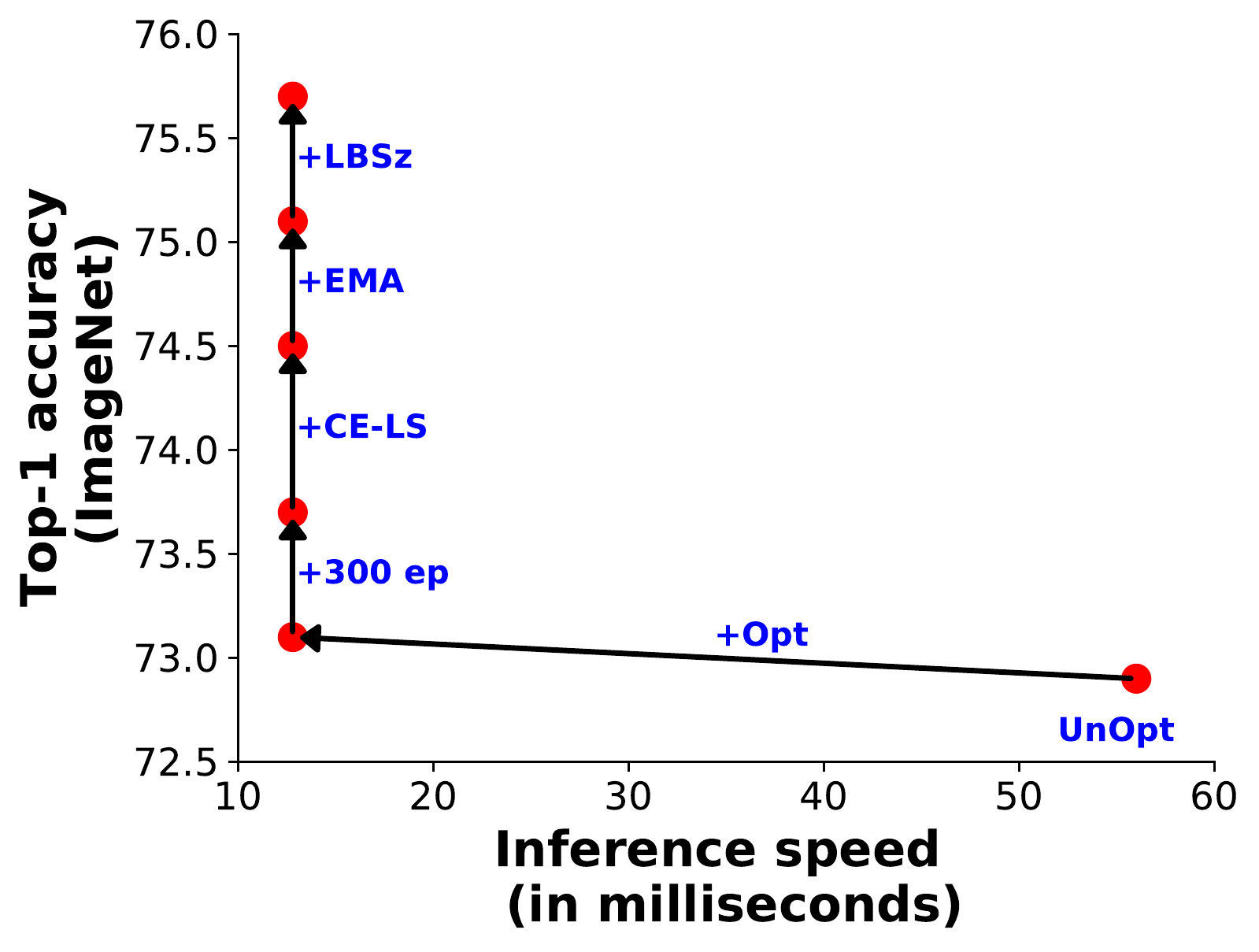}
    \caption{\textbf{Impact of different components in the training of \arch.} With \textbf{\textcolor{blue}{+}}, we indicate that the component is added to the previous configuration. Here, \textbf{\textcolor{blue}{UnOpt}} represents the un-optimized \arch~trained for 150 epochs with a batch size of 512 with cross entropy and \textbf{\textcolor{blue}{Opt}} represents \arch~with custom CUDA kernel. \textbf{\textcolor{blue}{300 ep}} denotes that model is trained for 300 epochs, \textbf{\textcolor{blue}{CE-LS}} denotes that label-smooth cross-entropy is used, \textbf{\textcolor{blue}{EMA}} denoted that exponential moving average is used, and \textbf{\textcolor{blue}{LBSz}} denotes that large batch size (2048 images) is used for training. Here, inference time is measured on NVIDIA GTX 1080 Ti GPU and is an average across 100 trials for a batch of 32 RGB images, each with a spatial dimension of $224\times 224$. }
    \label{fig:training_hyper}
\end{figure}

\subsection{Effect of MobileNet's training hyper-parameters}
\label{ssec:opt_hyper}
In previous experiments, we follow the experimental setup similar to ESPNetv2 and ShuffleNetv2 i.e., each model is trained for 150 epochs using a batch size of 512 and minimizes cross-entropy loss using SGD. Recent efficient models (e.g., MobileNetv3 and MNASNet) are trained longer (600 epochs) with extremely large batch size (4096) using cross-entropy with label smoothing (CE-LS) and exponential moving average (EMA). We also trained \arch~with CE-LS and EMA, with an exception to batch size (2048) and number of epochs (300). For training with larger batch size, we accumulated the gradients for 4 iterations. This resulted in an effective batch size of 2048 (128 images per NVIDIA GTX 1080 Ti GPU $\times$ 4 GPUs $\times$ accumulation frequency of 4). Figure \ref{fig:training_hyper} shows the effect of these changes. Similar to state-of-the-art models (e.g., MobileNetv2, MobileNetv3, MixNet, and MNASNet), \arch~also benefits from these hyper-parameters and yields a top-1 accuracy of 75.7 on the ImageNet. 

Importantly, the optimized CUDA kernel (Figure \ref{fig:cuda_implement}) improved the inference speed drastically over the unoptimized version. This is because the optimized kernel launches one kernel for these three branches compared to one per branch in the unoptimized one and also, maximizes work done per CUDA thread. This reduces latency. With optimized CUDA kernels, \arch~models (10-300 MFLOPs) takes between one and three days for training on the ImageNet dataset on 4 NVIDIA GTX 1080 Ti GPUs with an effective batch size of 2048.

\section{Evaluating \arch~on the ImageNet}
\label{sec:resultsImg}
In this section, we evaluate the performance of \arch~on the ImageNet dataset and show that \arch~delivers significantly better performance than state-of-the-art efficient networks, including neural search architectures. Recall that the \arch~is ShuffleNetv2 with the \dice~unit (Section \ref{ssec:archDetails}).

\noindent \textbf{Implementation details:} We scale the number of output channels by a width scaling factor $s$ to obtain \arch~models at different complexity levels, ranging from 6 MFLOPs to 500+ MFLOPs (see Appendix \ref{sec:append_arch} for details).

\vspace{1mm}
\noindent \textbf{Evaluation metrics and baselines:} We use $224\times 224$ single crop top-1 accuracy to evaluate the performance on the validation set. The performance of \arch~is compared with state-of-the-art \textit{manually} designed efficient networks (MobileNets \cite{howard2017mobilenets, sandler2018mobilenetv2}, ShuffleNetv2 \cite{ma2018shufflenet}, CondenseNet \cite{huang2017condensenet}, and ESPNetv2 \cite{mehta2018espnetv2}) and \textit{automatically} designed networks (MNASNet \cite{tan2018mnasnet}, FBNet \cite{wu2018fbnet}, MixNet \cite{Tan2019MixConvMD}, and MobileNetv3 \cite{howard2019searching}).

\vspace{1mm}
\noindent \textbf{Results:} Recent studies (e.g., MobileNetv3) have shown that longer training with extremely large batch sizes improves performance. To have fair comparisons with state-of-the-art methods, we report the performance of \arch~on two settings. The first setting, \arch-E150-B512, is similar to networks like ESPNetv2, ShuffleNetv2, and CondenseNet where \arch~is trained for fewer epochs (150) with a smaller batch size (512) without EMA and label smoothing. The second setting, \arch-E300-B2048, is similar to networks like MobileNets, MNASNet, and MixNet, where \arch~is trained for 300 epochs with a batch size of 2048. Table \ref{tab:imgnetRes} compares the performance of \arch~with state-of-the-art efficient architectures at different FLOP ranges.

Compared to networks that are trained with smaller batch sizes and fewer epochs, e.g., ESPNetv2 (epochs: 300; batch size: 512) and ShuffleNetv2 (epochs: 240 and batch size: 1024), \arch~delivers better performance across different FLOP ranges. Similarly, when \arch~is trained for longer with larger batch sizes, it delivered similar or better performance than state-of-the-art methods, including neural architecture search (NAS)-based methods. For about 300 MFLOPs, \arch~is 4\% more accurate than MobileNetv2. Specifically, we observe that \arch~is very effective when model size is small (FLOPs $<$ 150 M). For example, \arch~outperforms MNASNet \cite{tan2018mnasnet}, FBNet \cite{wu2018fbnet}, and MobileNetv3 \cite{howard2019searching} by 6.1\%, 3.2\%, and 1.1\% for network size of about 70 MFLOPs, respectively. 

Overall, these results shows that \dice~unit learns better representations than separable convolutions. We believe that incorporating the \dice~unit with NAS would yield better network.

\vspace{1mm}
\noindent \textbf{Inference speed:} We measure the inference time on two GPUs: (1) embedded or mobile GPU (NVIDIA GTX 960M) and (2) desktop GPU (NVIDIA GTX 1080 Ti)\footnote{We do not measure the inference speed on Smartphones because efficient implementations of \dice~unit are not yet available for such devices.}. \arch~is as fast as ShuffleNetv2 but delivers better performance. Compared to MobileNetv2 and MobileNetv3, \arch~has low latency while delivering similar or better performance. We observe that MobileNetv2 and MobileNetv3 models are slow in comparison to ShuffleNetv2 and \arch~when the batch size increases. This is because the number of channels in the depth-wise convolution in the inverted residual block of MobileNetv2 and MobileNetv3 are very large as compared to \arch~and ShuffleNetv2. For example, the maximum number of channels in depth-wise convolution in MobileNetv2 (300 MFLOPs) are 960 while the maximum number of channels in \dimconv~and depth-wise convolution in \arch~(298 MFLOPs) and ShuffleNetv2 (292 MFLOPs) are 576 and 352, respectively.

\section{Task-level Generalization of \arch}
\label{sec:results}
Several previous works (e.g., \cite{long2015fully, zhao2019object}) have shown that high accuracy on the Imagenet dataset does not necessarily correlates with high accuracy on visual scene understanding tasks (e.g., object detection and semantic segmentation). Since these tasks are widely used in real-world applications (e.g., autonomous wheel chair and robots) and often run on resource-constrained devices (e.g., embedded devices), it is important that efficient networks generalizes well on these tasks. Therefore, we evaluate the performance of \arch~on three different tasks: (1) object detection (Section \ref{ssec:objDec}), (2) semantic segmentation (Section \ref{ssec:semSeg}), and (3) multi-object classification (Section \ref{ssec:multObj}). Compared to existing efficient networks that are built using separable convolutions (e.g., MobileNets \cite{howard2017mobilenets, sandler2018mobilenetv2, howard2019searching}, MixNet \cite{Tan2019MixConvMD}, and ESPNetv2 \cite{mehta2018espnetv2}), \arch~delivers better performance. 

\subsection{Object Detection on VOC and MS-COCO}
\label{ssec:objDec}

\vspace{1mm}
\noindent \textbf{Implementation details:} For object detection, we use Single Shot object Detection (SSD) \cite{liu2016ssd} pipeline. We use \arch~(298 MFLOPs) pretrained on the ImageNet as a base feature extractor instead of VGG \cite{simonyan2014very}. We fine-tune our network using SGD with smooth L1 and cross-entropy losses for object localization and classification, respectively.

\vspace{1mm}
\noindent \textbf{Evaluation metrics and baselines:} We evaluate the performance using mean Average Precision (mAP). For MS-COCO, we report mAP@IoU of 0.50:0.95. For SSD as a detection pipeline, we compare \arch's performance with two types of base feature extractors: (1) \emph{manual} (VGG \cite{simonyan2014very}, MobileNet \cite{howard2017mobilenets}, MobileNetv2 \cite{sandler2018mobilenetv2}, and ESPNetv2 \cite{mehta2018espnetv2}) and (2) \emph{NAS-based} (MNASNet \cite{tan2018mnasnet}, MixNet \cite{Tan2019MixConvMD}, and MobileNetv3 \cite{howard2019searching}).

\vspace{1mm}
\noindent \textbf{Results:} Table \ref{tab:objDet} compares quantitative results of SSD with different backbone networks on the PASCAL VOC 2007 and the MS-COCO datasets. \arch~significantly improves the performance of SSD-based object detection pipeline and delivers 1-4\% higher mAP than other existing efficient variants of SSD, including NAS-based backbones such as MobileNetv3 (22.0 vs. \textbf{25.1}) and MixNet (22.3 vs. \textbf{25.1}). Compared to standard SSD with VGG as backbone, \arch~achieves higher mAP while being $\mathbf{45\times}$ and $\mathbf{38\times}$ more efficient on the PASCAL VOC 2007 and the MS-COCO dataset, respectively. 

\begin{table}[t!]
    \centering
        \resizebox{0.95\columnwidth}{!}{
        \begin{tabular}{lcrrrrr}
            \toprule[1.5pt]
            \multirow{2}{*}{\textbf{SSD backbone}} & \textbf{Image}  & \multicolumn{2}{c}{\textbf{VOC07}} & & \multicolumn{2}{c}{\textbf{MS-COCO}} \\
            \cmidrule[1.25pt]{3-4} \cmidrule[1.25pt]{6-7}
             & \textbf{size} & \textbf{FLOPs} & \textbf{mAP} & & \textbf{FLOPs} & \textbf{mAP} \\
            \midrule[1pt]
            \multirow{2}{*}{VGG \cite{simonyan2014very}} & 512x512  & 90.2 B & 74.9 & & 99.5 B & 26.8 \\
              & 300x300 & 31.3 B & 72.4 & & 35.2 B & 23.2 \\
            \midrule
            MobileNet \cite{howard2017mobilenets} & 320x320 & -- & --  && 1.3 B & 22.2 \\
            MobileNetv2 \cite{sandler2018mobilenetv2} & 320x320 & -- & --  && \textbf{0.8 B} & 22.1 \\
            \midrule
            \multirow{2}{*}{ESPNetv2 \cite{mehta2018espnetv2}} & 512x512 & 2.5 B & 75.0  & & 3.2 B & 26.0 \\
            & 256x256 & 0.9 B & 70.3  & & 1.1 B & 21.9 \\
            \midrule
            \arch~(Ours) & 512x512 & 2.0 B & \textbf{77.2}  & & 2.6 B & \textbf{28.0} \\
            \arch~(Ours) & 300x300 & \textbf{0.7 B} & 71.9 & & 0.9 B & 25.1 \\
            \midrule
            \midrule
            MobileNetv3 \cite{howard2019searching} (NAS) & -- & -- & -- & & 0.6 B & 22.0 \\
            MixNet \cite{Tan2019MixConvMD} (NAS) & -- & -- & -- & & 0.9 B & 22.3 \\
            MNASNet \cite{tan2018mnasnet} (NAS) & -- & -- & -- & & 0.8 B & 23.0 \\
            \bottomrule[1.5pt]
        \end{tabular}
        }
    \caption{\textbf{Object detection} results of SSD \cite{liu2016ssd} with different backbones on PASCAL VOC 2007 and MS-COCO. On the MS-COCO dataset, total network parameters in SSD with different backbones are about 5 million, except VGG. See Appendix \ref{sec:append_quant} for qualitative results.}
    \label{tab:objDet}
\end{table}

\begin{figure}[t!]
    \centering
    \includegraphics[width=0.8\columnwidth]{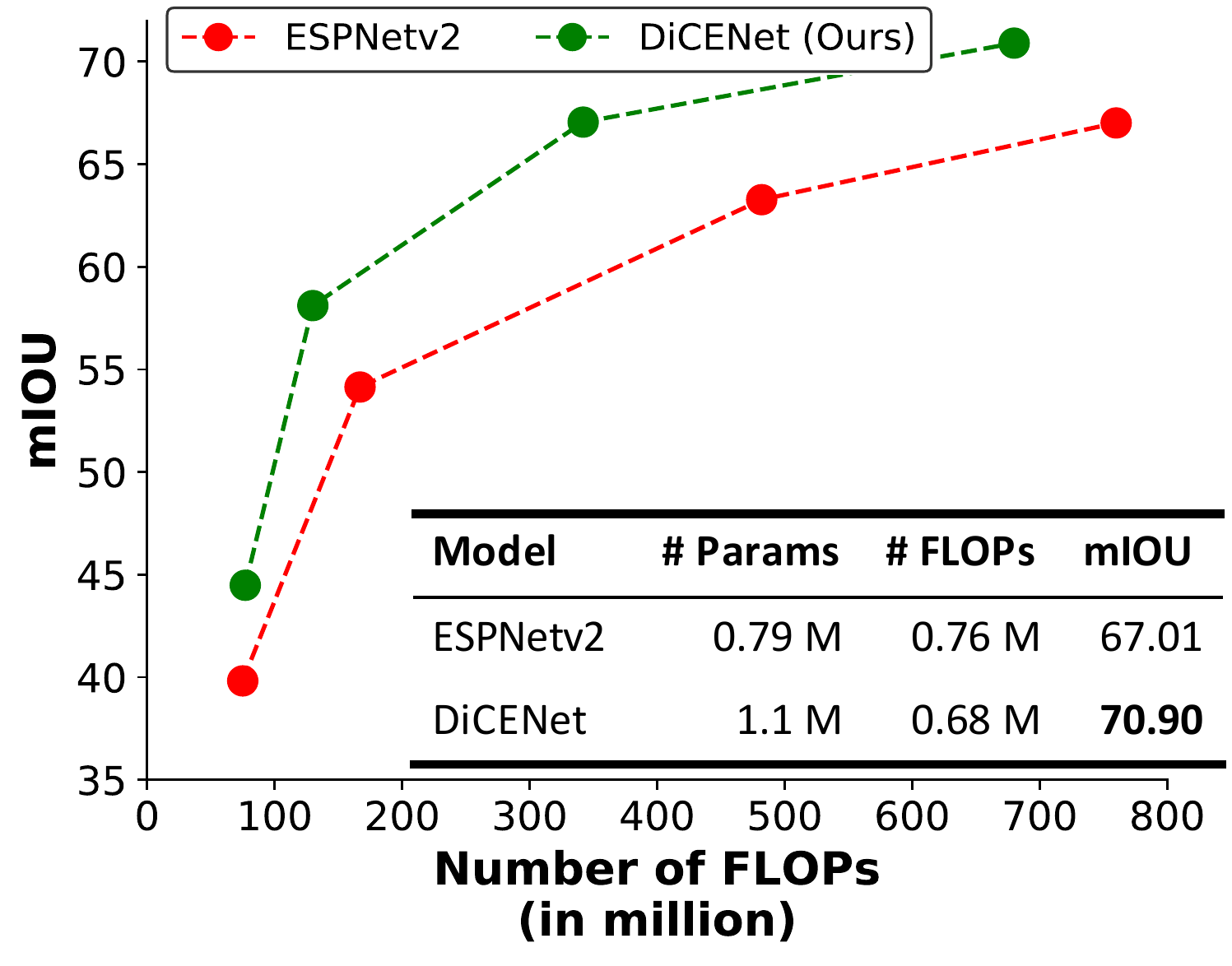}
    \caption{\textbf{Semantic segmentation} results on the PASCAL VOC 2012 validation set. Here, mIOU represents mean intersection over union. See Appendix \ref{sec:append_quant} for qualitative results.}
    \label{fig:semseg}
\end{figure}

\subsection{Semantic Segmentation on PASCAL VOC}
\label{ssec:semSeg}

\vspace{1mm}
\noindent \textbf{Implementation details:} We adapt \arch~to ESPNetv2's \cite{mehta2018espnetv2} encoder-decoder architecture. We choose this network because it delivers competitive performance to existing methods even with low resolution images (e.g. 256$\times$256 vs. 512$\times$512). We replace the encoder in ESPNetv2 (pretrained on ImageNet) with the \arch~and follow the same training procedure for fine-tuning as ESPNetv2. We do not change the decoder.

\vspace{1mm}
\noindent \textbf{Evaluation metrics and baselines:} The performance at different complexity levels (FLOPs) is evaluated using mean intersection over union (mIOU).

\vspace{1mm}
\noindent \textbf{Results:} Figure \ref{fig:semseg} compares the performance of \arch~with ESPNetv2 on the PASCAL VOC 2012 validation set. \arch~significantly improves the segmentation performance i.e., for similar FLOPs, \arch~achieves higher mIOU while for similar mIOU, \arch~requires significantly fewer FLOPs.

\begin{table}[t!]
    \centering
    \resizebox{0.85\columnwidth}{!}{
        \begin{tabular}{lcccc}
            \toprule[1.5pt]
             \multirow{2}{*}{\textbf{Network}} & \multirow{2}{*}{\textbf{\# Params}} & \multirow{2}{*}{\textbf{\# FLOPs}} & \multicolumn{2}{c}{\textbf{F1-score}} \\
             \cmidrule[1.25pt]{4-5}
             & & & \textbf{Class-wise} & \textbf{Overall} \\
             \midrule[1pt]
             ShuffleNetv2 \cite{ma2018shufflenet}$^\dagger$ & 3.5 M & 300 M & 60.42 & 67.58 \\
             ESPNetv2 \cite{mehta2018espnetv2}$^\dagger$ & 3.5 M & 284 M & 63.41 & 69.23 \\
             \arch~(ours) & 5.1 M & 298 M  & \textbf{66.92} & \textbf{73.41} \\
             \bottomrule[1.5pt]
        \end{tabular}
    }
    \caption{\textbf{Multi-object classification} results on the MS-COCO dataset. \small{Here, $^\dagger$ indicates that results are from \cite{mehta2018espnetv2}.}}
    \label{tab:multiLabel}
\end{table}

\subsection{Multi-object Classification on MS-COCO}
\label{ssec:multObj}
\noindent \textbf{Implementation details:} Following \cite{mehta2018espnetv2}, we fine-tune \arch~using the binary cross-entropy loss.

\vspace{1mm}
\noindent \textbf{Evaluation metrics and baselines:} Similar to \cite{mehta2018espnetv2}, we evaluate the performance using overall and per-class F1 score and compare with two efficient architectures, i.e., ESPNetv2 and ShuffleNetv2.

\vspace{1mm}
\noindent \textbf{Results:} Table \ref{tab:multiLabel} shows that \arch~outperforms existing efficient networks by a significant margin (e.g., ESPNetv2 and ShuffleNetv2 by 4.1\% and 5.8\% respectively) on this task.

\section{Conclusion}
We introduce a novel and generic convolutional unit, the \dice~unit, that uses dimension-wise convolutions and dimension-wise fusion module to learn spatial and channel-wise representations efficiently. Our empirical results suggest that the \dice~unit is more effective than separable convolutions. Moreover, when we stack \dice~units to build \arch~model, we observe significant improvements across different computer vision tasks. We have shown that the \dice~unit is effective. Future work involves adding the \dice~unit in neural search space to discover a better neural architecture, particularly with \cite{tan2018mnasnet,Tan2019MixConvMD, howard2019searching}.

\ifCLASSOPTIONcompsoc
  \section*{Acknowledgments}
\else
  \section*{Acknowledgment}
\fi

This research was supported by ONR N00014-18-1-2826, DARPA N66001-19-2-403, NSF (IIS-1616112, IIS1252835), an Allen Distinguished Investigator Award, Samsung GRO and gifts from Allen Institute for AI, Google, and Amazon. Authors would also like to thank members of the PRIOR team at the Allen Institute for Artificial Intelligence (AI2), the H2Lab at the University of Washington, Seattle, and anonymous reviewers for their valuable feedback and comments.

\ifCLASSOPTIONcaptionsoff
  \newpage
\fi

\bibliographystyle{IEEEtran}
\bibliography{IEEEabrv,main.bib}

\appendices

\section{Network Architecture}
\label{sec:append_arch}
The overall architecture at different network complexities is given in Table \ref{tab:networkSup}. The first layer is a standard $3\times3$ convolution with a stride of two while the second layer is a max pooling layer. All convolutional layers are followed by a batch normalization layer \cite{ioffe2015batch} and a PReLU non-linear activation layer \cite{he2015delving}, except for the last layer that feeds into a softmax for classification. Following previous work \cite{howard2017mobilenets, sandler2018mobilenetv2, ma2018shufflenet, zhang2017shufflenet, mehta2018espnetv2}, we scale the number of output channels by a width scaling factor $s$ to construct networks at different FLOPs. We initialize weights of our network using the same method as in \cite{he2015delving}.

\section{Qualitative Results for Object Detection and Semantic Segmentation}
\label{sec:append_quant}
\begin{figure*}[t!]
    \centering
    \begin{subfigure}[b]{2\columnwidth}
        \centering
        \includegraphics[height=100px]{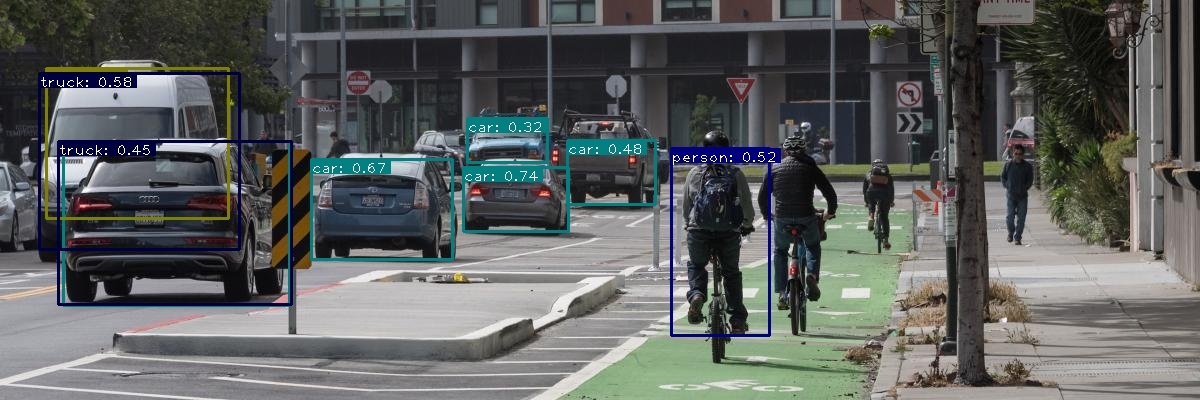}
        \includegraphics[height=100px]{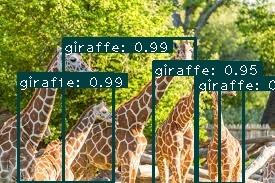}
    \end{subfigure}
    \vfill
    \begin{subfigure}[b]{2\columnwidth}
        \centering
        \includegraphics[height=110px]{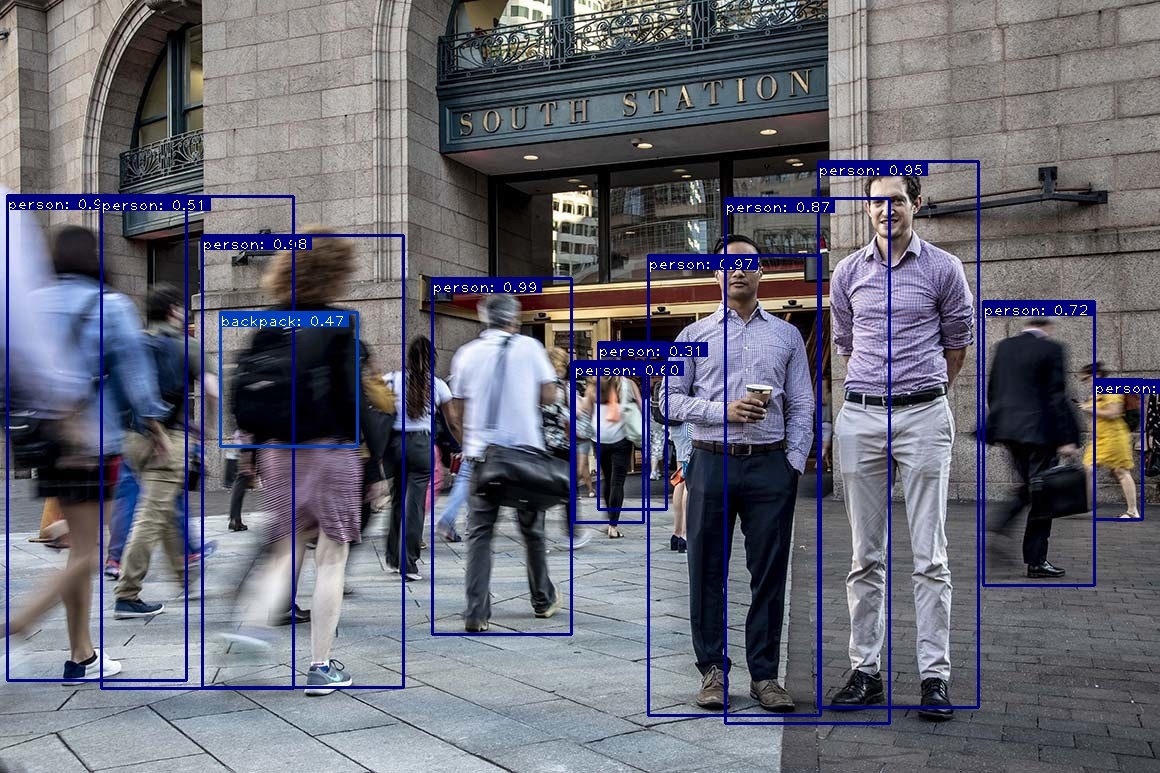}
        \includegraphics[height=110px]{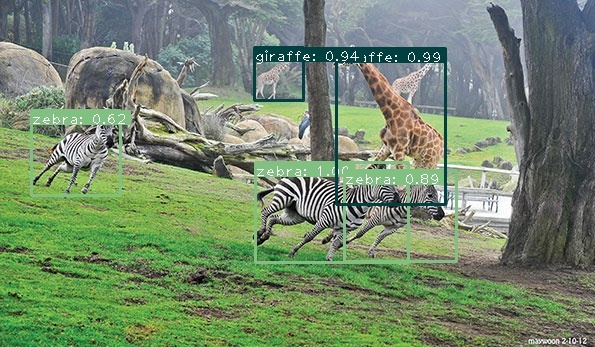}
        \includegraphics[height=110px]{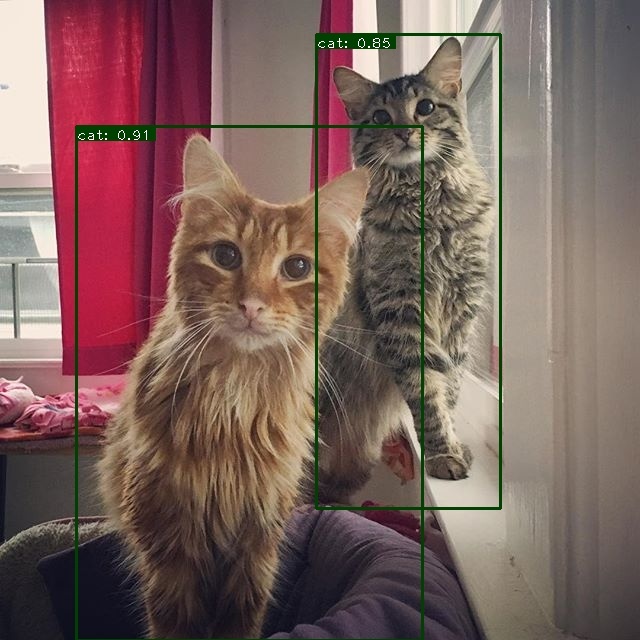}
    \end{subfigure}
    \vfill
    \begin{subfigure}[b]{2\columnwidth}
        \centering
        \includegraphics[height=95px]{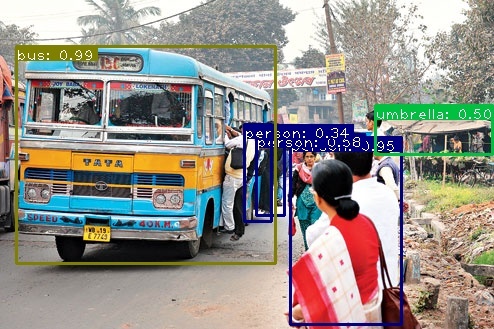}
        \includegraphics[height=95px]{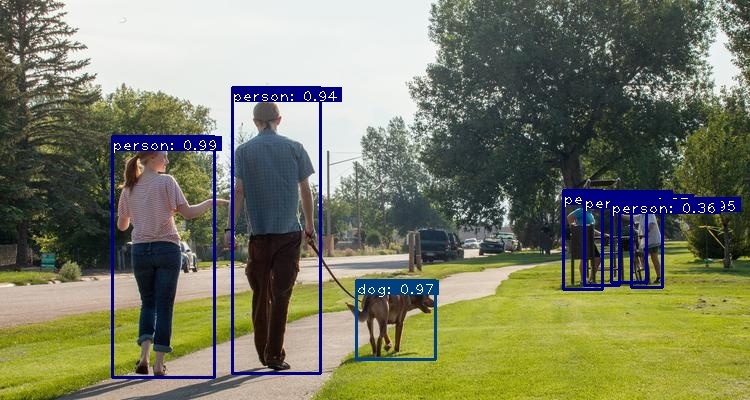}
        \includegraphics[height=95px]{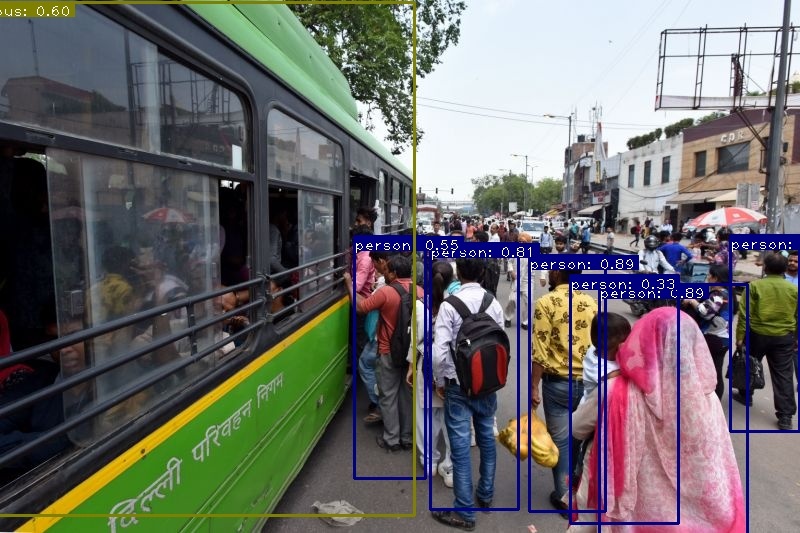}
    \end{subfigure}
    \vfill
    \begin{subfigure}[b]{2\columnwidth}
        \centering
        \includegraphics[height=86px]{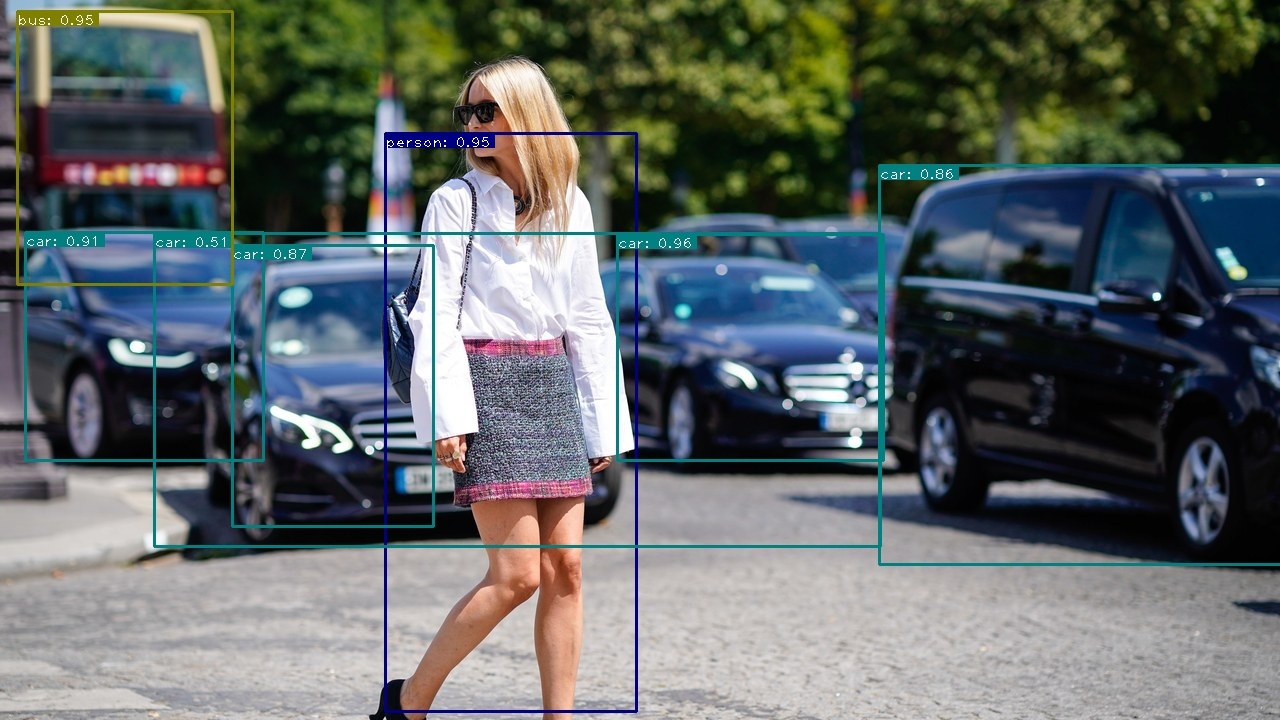}
        \includegraphics[height=86px]{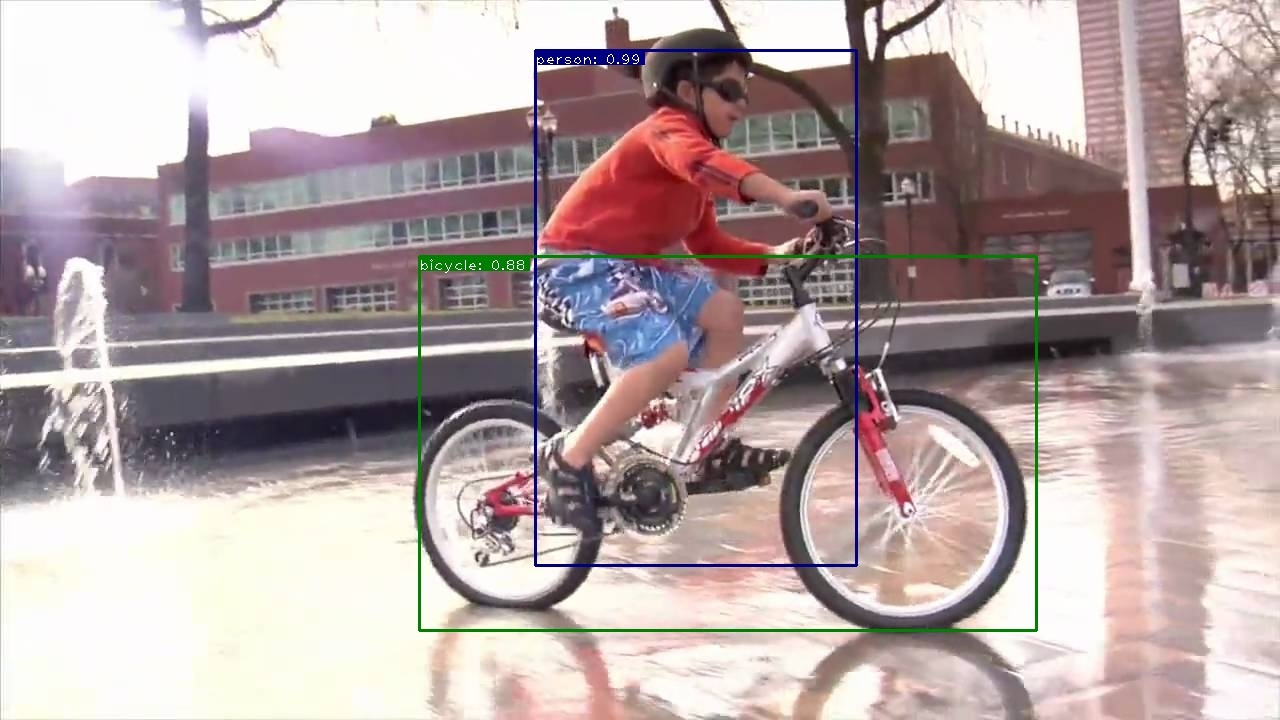}
        \includegraphics[height=86px]{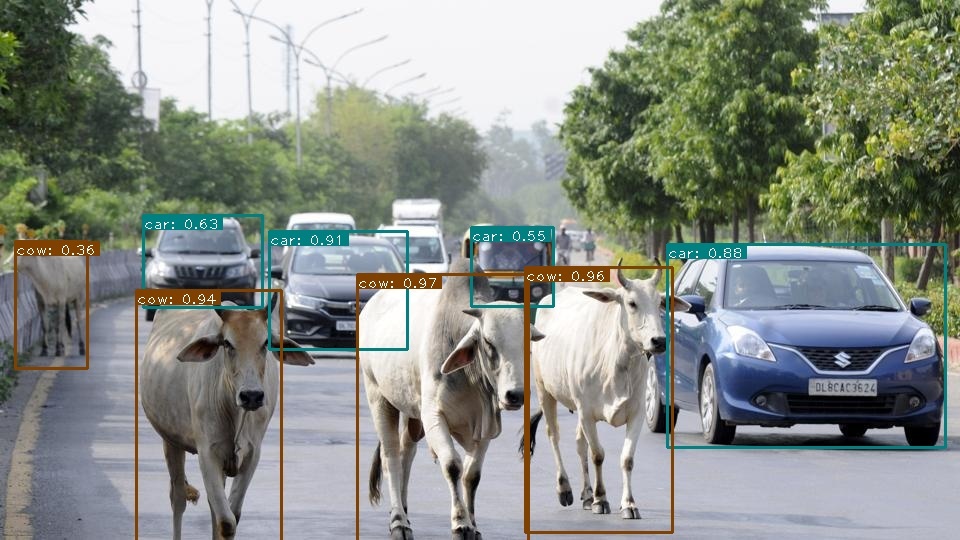}
    \end{subfigure}
    \caption{Object detection using \arch~under diverse background conditions.}
    \label{fig:objDet}
\end{figure*}

\begin{figure*}[t!]
    \centering
    \begin{subfigure}[b]{\columnwidth}
        \centering
        \includegraphics[width=0.8\columnwidth]{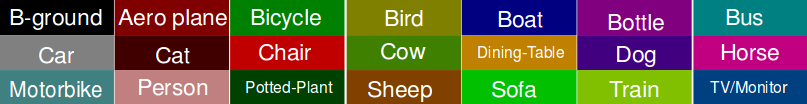}
        \caption{PASCAL VOC Colormap}
    \end{subfigure}
    \vfill
    \begin{subfigure}[b]{2\columnwidth}
        \centering
        \resizebox{\columnwidth}{!}{
                \begin{tabular}{cccccc}
                \includegraphics[height=50px]{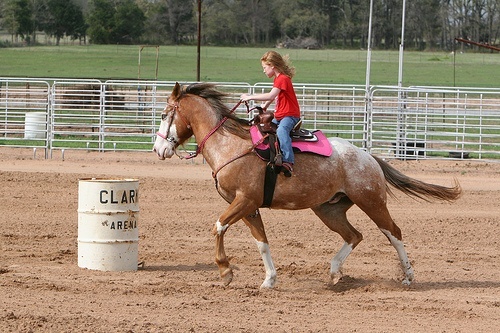} &
                \includegraphics[height=50px]{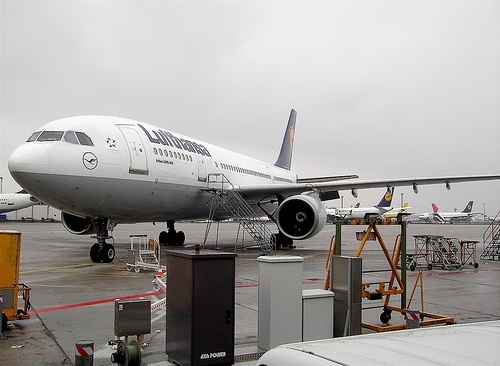} &
                \includegraphics[height=50px]{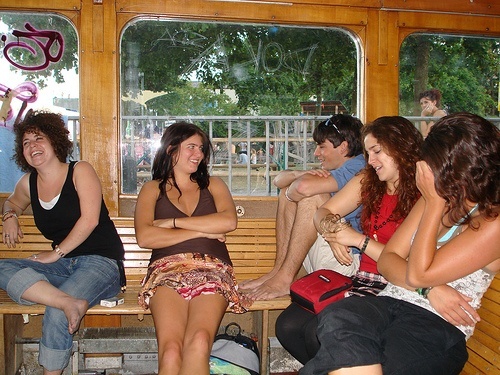} &
                 \includegraphics[height=50px]{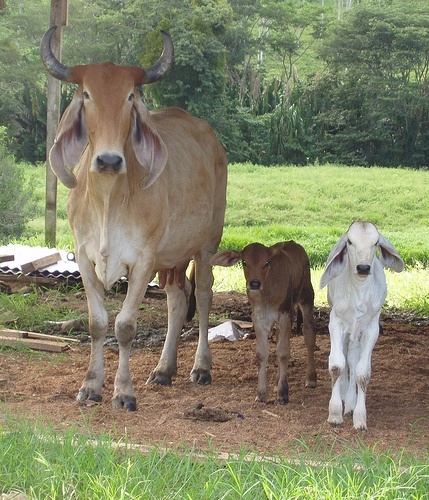} &
                 \includegraphics[height=50px]{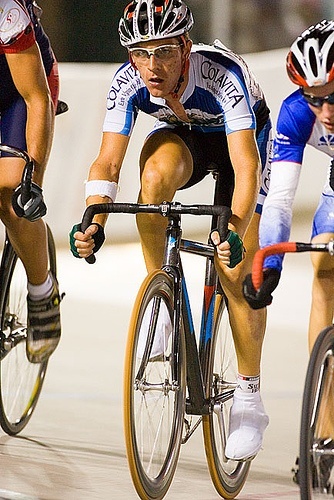} &
                 \includegraphics[height=50px]{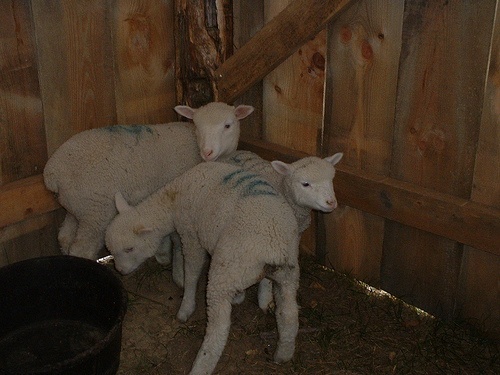} \\

                \includegraphics[height=50px]{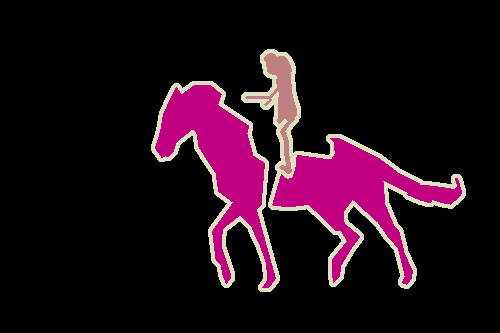} &
                \includegraphics[height=50px]{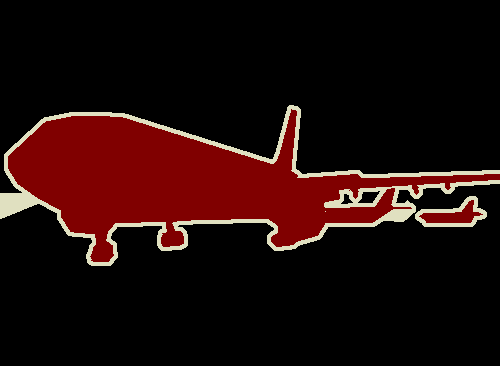} &
                \includegraphics[height=50px]{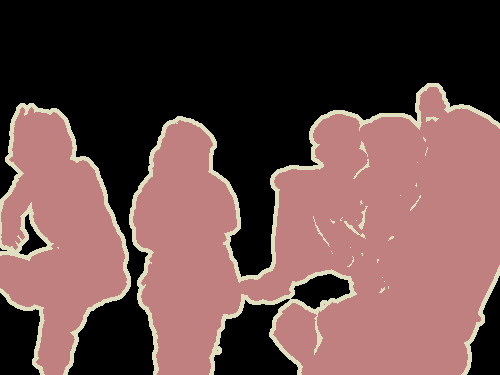}  &
                \includegraphics[height=50px]{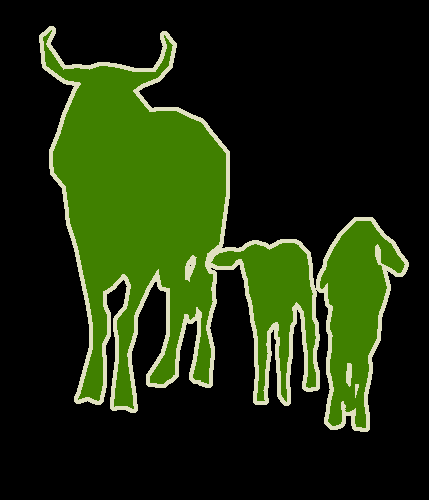} &
                \includegraphics[height=50px]{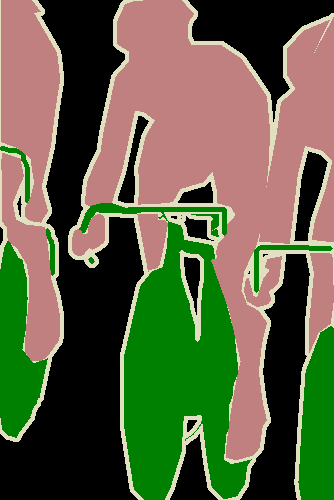} &
                \includegraphics[height=50px]{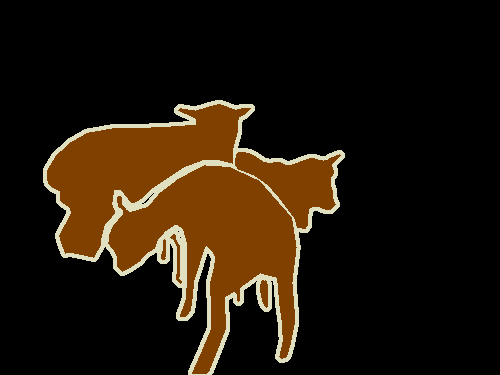} \\
                \includegraphics[height=50px]{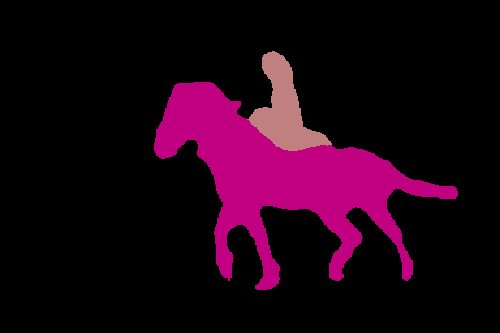} &
                \includegraphics[height=50px]{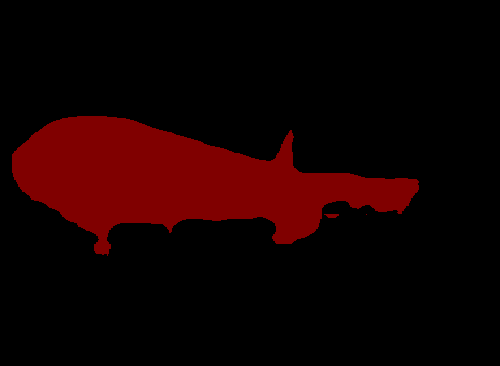} &
                \includegraphics[height=50px]{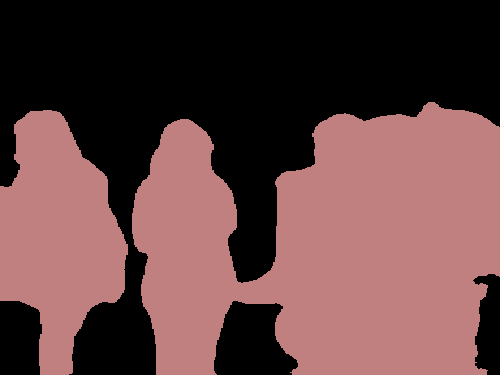} &
                \includegraphics[height=50px]{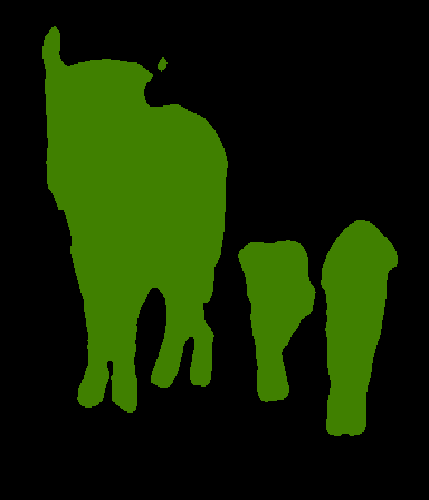} &
                \includegraphics[height=50px]{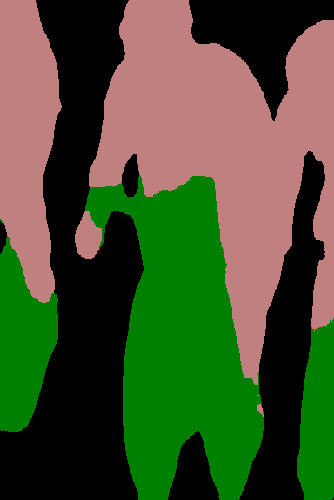} &
                \includegraphics[height=50px]{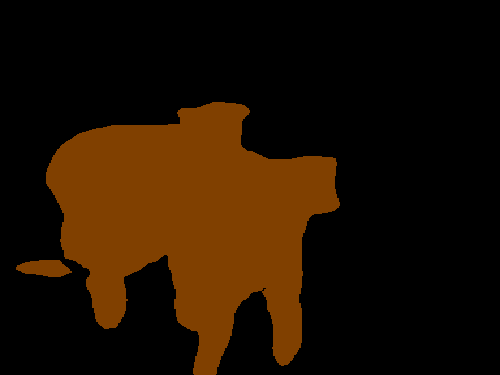} \\
                \end{tabular}
        }
        \caption{Results on the PASCAL VOC 2012 validation set (\textbf{top row}: Input image, \textbf{middle row}: ground truth, \textbf{last row}: \arch~predictions)}
     \end{subfigure}
     \vfill
    \begin{subfigure}[b]{2\columnwidth}
    \centering
        \resizebox{\columnwidth}{!}{
                \begin{tabular}{ccccc}
                \includegraphics[height=50px]{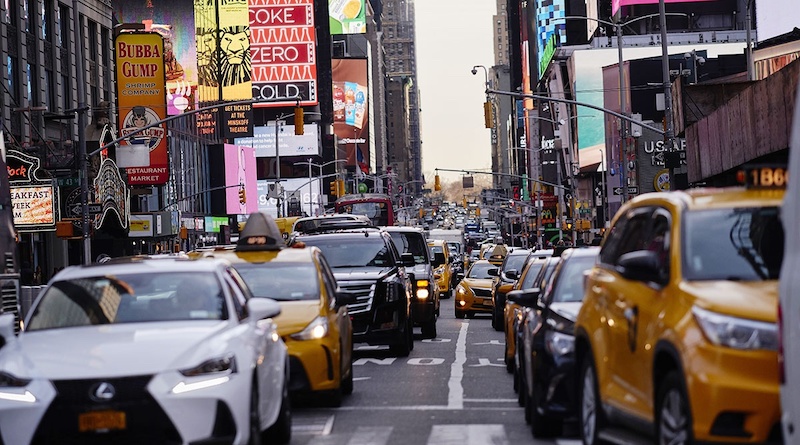} &
                \includegraphics[height=50px]{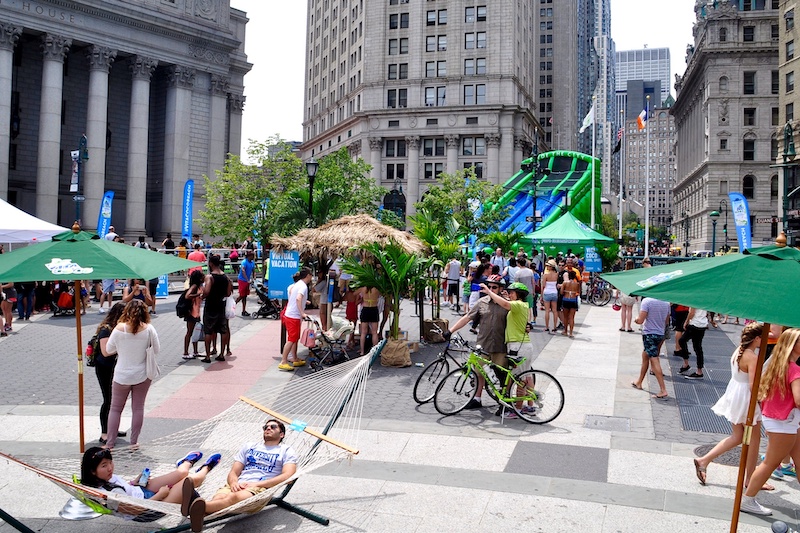} &
                \includegraphics[height=50px]{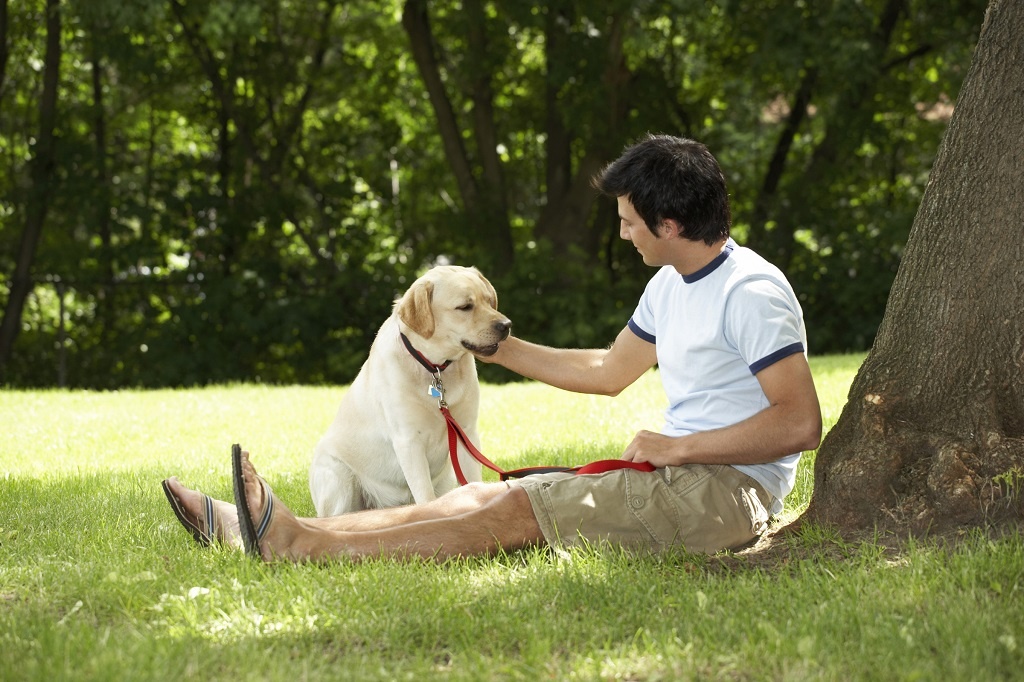} &
                \includegraphics[height=50px]{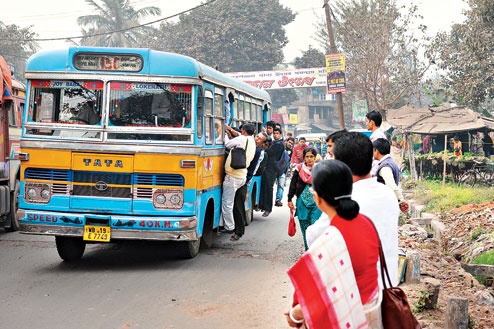} &
                \includegraphics[height=50px]{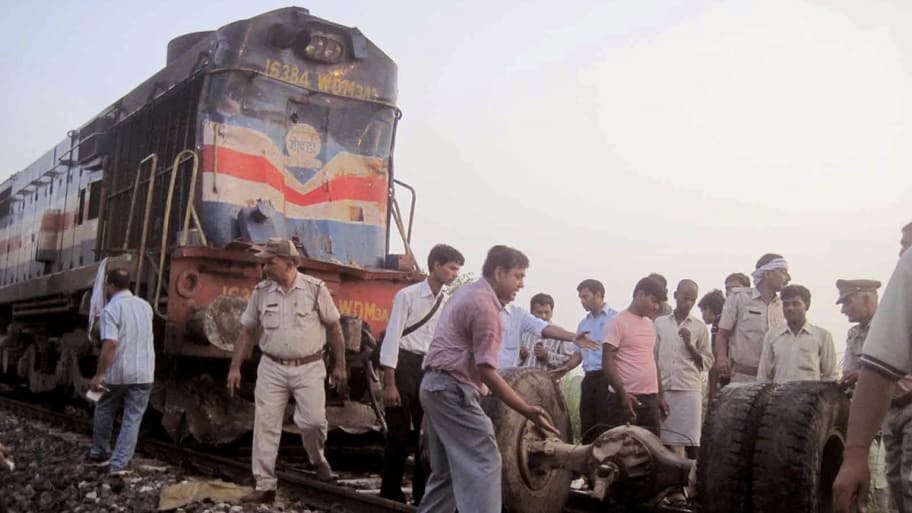} \\
                \includegraphics[height=50px]{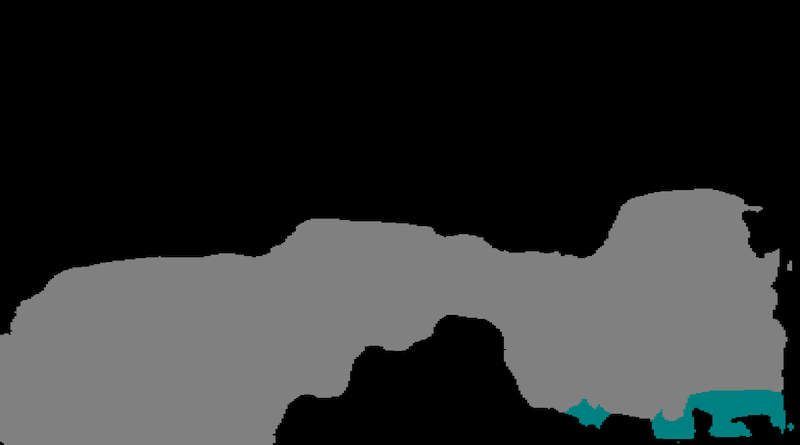} &
                \includegraphics[height=50px]{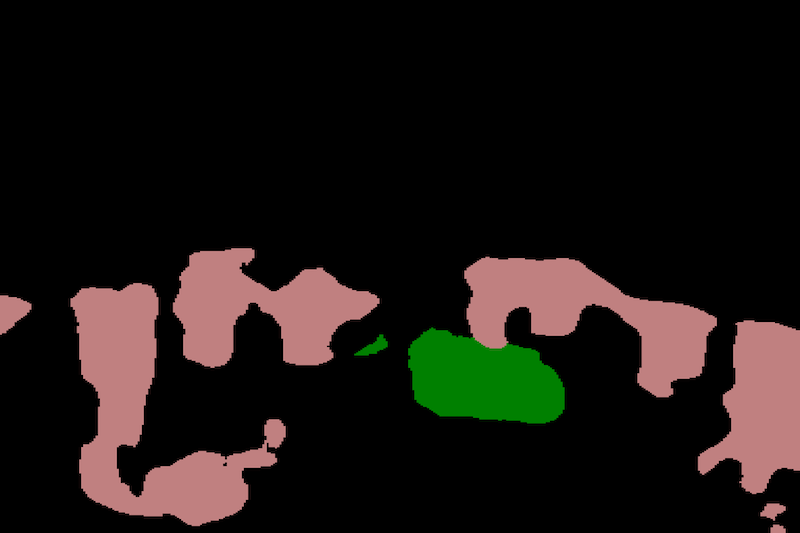} &
                \includegraphics[height=50px]{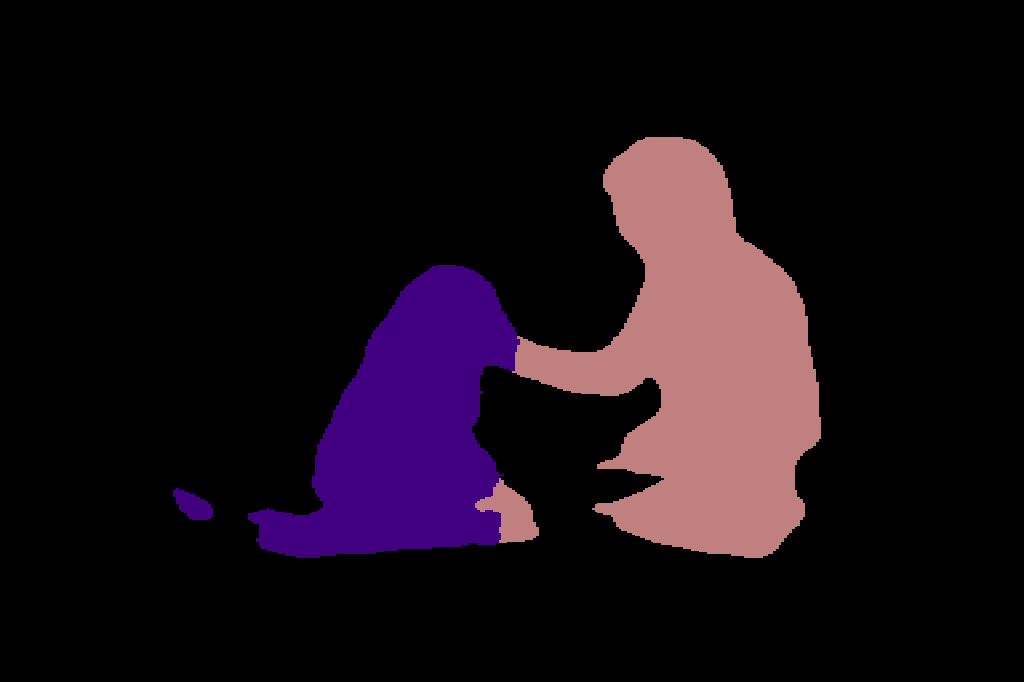} &
                \includegraphics[height=50px]{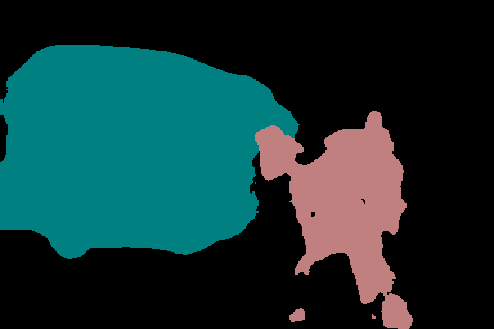} &
                \includegraphics[height=50px]{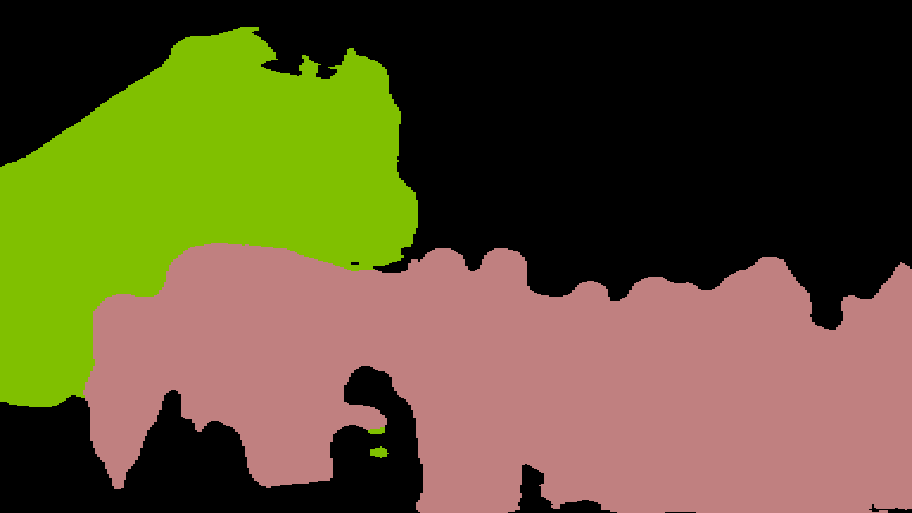} \\
                \includegraphics[height=50px]{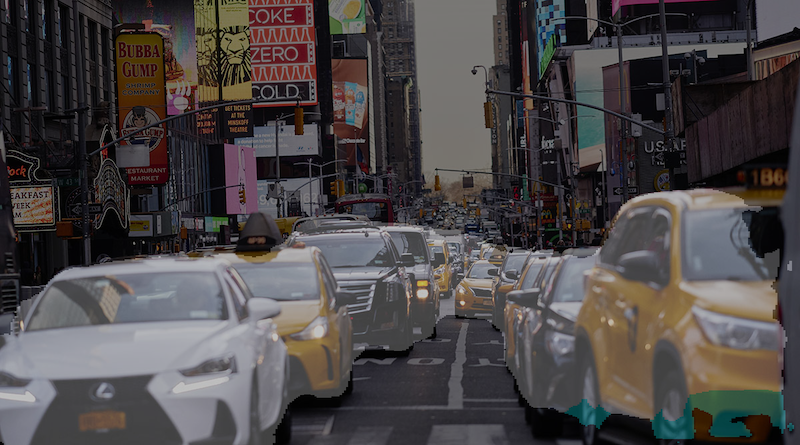} &
                \includegraphics[height=50px]{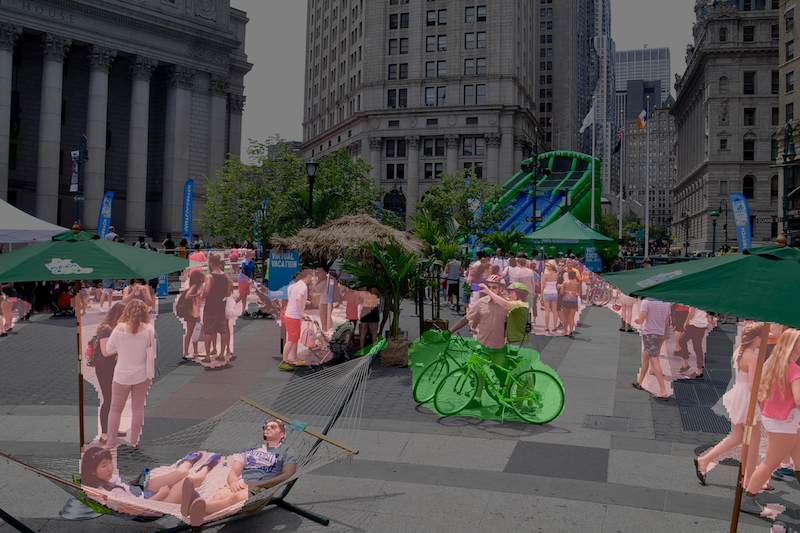} &
                \includegraphics[height=50px]{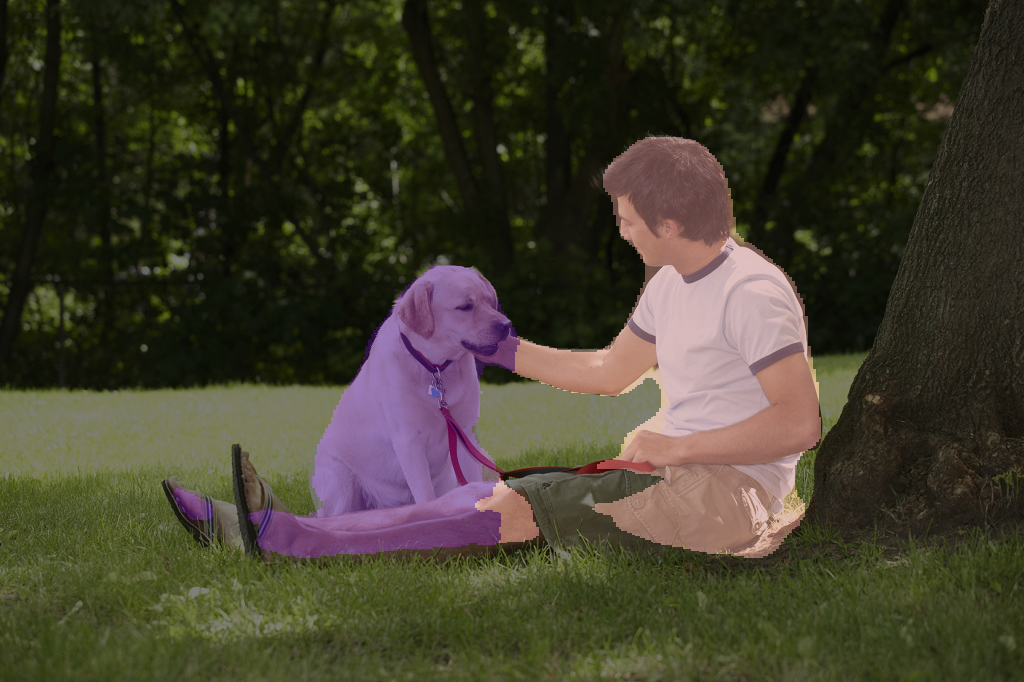} &
                \includegraphics[height=50px]{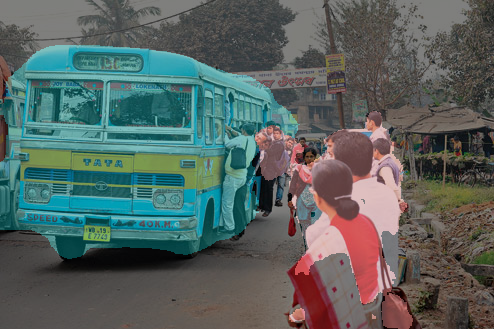} &
                \includegraphics[height=50px]{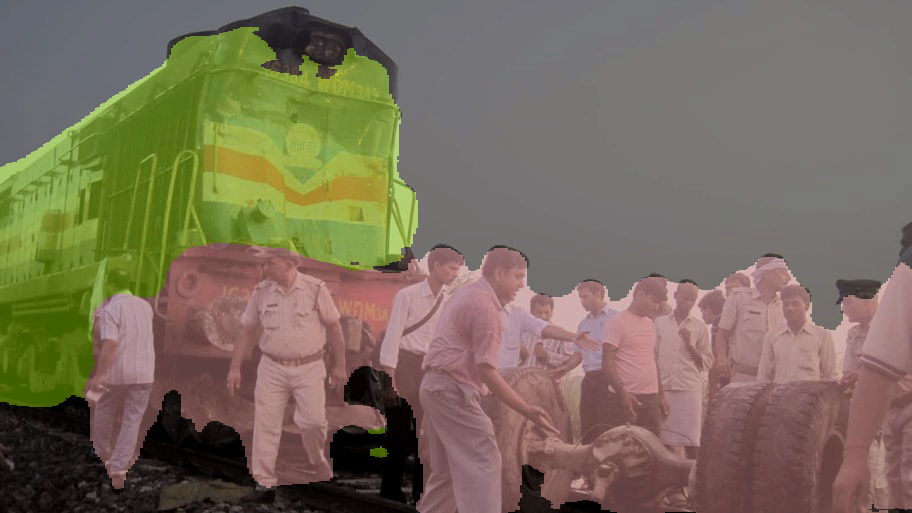} \\
                \end{tabular}
        }
        \caption{Results in the wild (\textbf{top row}: Input image, \textbf{middle row}: predictions, \textbf{last row}: predictions overlayed on the input image). These images do not have ground truth labels. Therefore, to reflect the segmentation quality, we overlayed segmentation masks on the input images.}
     \end{subfigure}
    \caption{Semantic segmentation results}
    \label{fig:semsegwild}
\end{figure*}

Figure \ref{fig:objDet} provides qualitative results for object detection while Figures \ref{fig:semsegwild} provide results for semantic segmentation on the PASCAL VOC 2012 dataset as well as \textit{in the wild}. These results suggest that the \arch~is able to detect and segment objects in diverse settings, including different backgrounds and image sizes.

\begin{table*}[t!]
    \centering
    \resizebox{1.5\columnwidth}{!}{
        \begin{tabular}{lcccccccc}
        \toprule[1.5pt]
            \multirow{2}{*}{\textbf{Layer}} &  \multirow{2}{*}{\textbf{Output size}} & \multirow{2}{*}{\textbf{Kernel size}} & \multirow{2}{*}{\textbf{Stride}}  & \multirow{2}{*}{\textbf{Repeat}} & \multicolumn{4}{c}{\bf Output channels (network width scaling parameter $s$)}\\
            \cmidrule[1.25pt]{6-9}
            & & & & & $s=0.1$ & $s=0.2$ & $s \in \left[0.5, 2.0\right]$ & $s=2.4$\\
            \midrule[1pt]
            Image & $224 \times 224$ & & & & 3 & 3 & 3 & 3 \\
            \midrule
            Conv1 & $112 \times 112$ & $3 \times 3$ & 2 & 1 & 8 & 16 & 24 & 24 \\
            Max Pool & $56 \times 56$ & $3 \times 3$ & 2 & & 8 & 16 & 24 & 24 \\
            \midrule
            \multirow{2}{*}{Block 1} & $28 \times 28$ &  & 2 & 1 & 16 & 32 & 116 $\times s$ & 278 \\
             & $28 \times 28$ &  & 1 & 3 & 16 & 32 & 116 $\times s$ & 278  \\
             \midrule 
             \multirow{2}{*}{Block 2} & $14 \times 14$ &  & 2 & 1 & 32 & 64 & 232 $\times s$ & 556 \\
             & $14 \times 14$ &  & 1 & 7 & 32 & 64 & 232 $\times s$ & 556 \\
             \midrule 
             \multirow{2}{*}{Block 3} & $7 \times 7$ &  & 2 & 1 & 64 & 128 & 464 $\times s$ & 1112 \\
             & $7 \times 7$ &  & 1 & 3 & 64 & 128 & 464 $\times s$ & 1112 \\
             \midrule
             Global Pool & $1\times 1$ & $7 \times 7$ & \multicolumn{2}{c}{} & 512 & 1024 & 1024 & 1280\\
             \midrule
             Grouped FC \cite{mehta2018pru} & $1 \times 1$ & $1\times1$ & 1 & 1 & 512 & 1024 & 1024 & 1280 \\
             \midrule
             FC & \multicolumn{4}{c}{ } & 1000 & 1000 & 1000 & 1000 \\
             \midrule[1pt]
             FLOPs & \multicolumn{4}{c}{ } & 6.5 M & 12 M & 24-240 M & 298 M \\
        \bottomrule[1.5pt]
        \end{tabular}
    }
    \caption{Overall architecture of \arch~(ShuffleNetv2 with \dice~unit) at different network complexities for the ImageNet classification. We use 4 groups in grouped fully connected (FC) layer. For other architecture designs in Figure 5 on the paper (MobileNetv1 and ResNet), we replace blocks 1, 2, and 3 with the corresponding blocks.}
    \label{tab:networkSup}
\end{table*}

\end{document}